%% file: FullArXivPaper.tex
\pdfoutput=1
\documentclass[10pt,twocolumn,letterpaper]{article}

\usepackage[pagenumbers]{cvpr} 

\makeatletter
\@namedef{ver@everyshi.sty}{}
\makeatother

\usepackage{graphicx}
\usepackage{amsmath}
\usepackage{amssymb}
\usepackage{booktabs}
\usepackage{flushend}

\usepackage{todonotes}
\usepackage{outlines}
\usepackage{caption}
\usepackage{subcaption}
\usepackage{microtype}
\usepackage{hhline}
\usepackage{multirow}
\usepackage{siunitx}
\usepackage{makecell}
\usepackage{gensymb}
\usepackage[inline]{enumitem}
\usepackage{placeins}

\usepackage{algorithm}
\usepackage{algpseudocode}
\usepackage{cuted}
\usepackage[accsupp]{axessibility}  

\usepackage{ifthen}
\newboolean{reftosupp}
\setboolean{reftosupp}{true}
\newboolean{reftomain}
\setboolean{reftomain}{true}

\usepackage{pifont}
\newcommand{\cmark}{\ding{51}}%
\newcommand{\xmark}{\ding{55}}%

\usepackage{hyperref}
\hypersetup{pagebackref,breaklinks,colorlinks}

\usepackage[capitalize]{cleveref}
\crefname{section}{Sec.}{Secs.}
\Crefname{section}{Section}{Sections}
\Crefname{table}{Table}{Tables}
\crefname{table}{Tab.}{Tabs.}
\crefname{algorithm}{Algo.}{Algos.}
\Crefname{algorithm}{Algorithm}{Algorithms}
\crefname{appendix}{Sec.}{Secs.}
\Crefname{appendix}{Section}{Sections}

\input{content/assets/variable_definitions}

\def\cvprPaperID{9001} 
\def\confName{CVPR}
\def\confYear{2023}

\title{\ourName: \ourNameFull}

\author{
Nick~Heppert$^{1,3}$,~
Muhammad~Zubair~Irshad$^{2}$,~
Sergey~Zakharov$^{3}$,~
Katherine~Liu$^{3}$,\\
Rares~Andrei~Ambrus$^{3}$,~
Jeannette~Bohg$^{4}$,~
Abhinav~Valada$^{1}$,~
Thomas~Kollar$^{3}$\vspace{2mm}\\
$^{1}$University~of~Freiburg\quad$^{2}$Georgia~Institute~of~Technology\\
$^{3}$Toyota~Research~Institute~(TRI)\quad$^{4}$Stanford~University
}

\begin{document}
\twocolumn[{%
\renewcommand\twocolumn[1][]{#1}%
\maketitle
\vspace{-6mm}
\begin{center}
    \centering
    \includegraphics[width=0.98\textwidth]{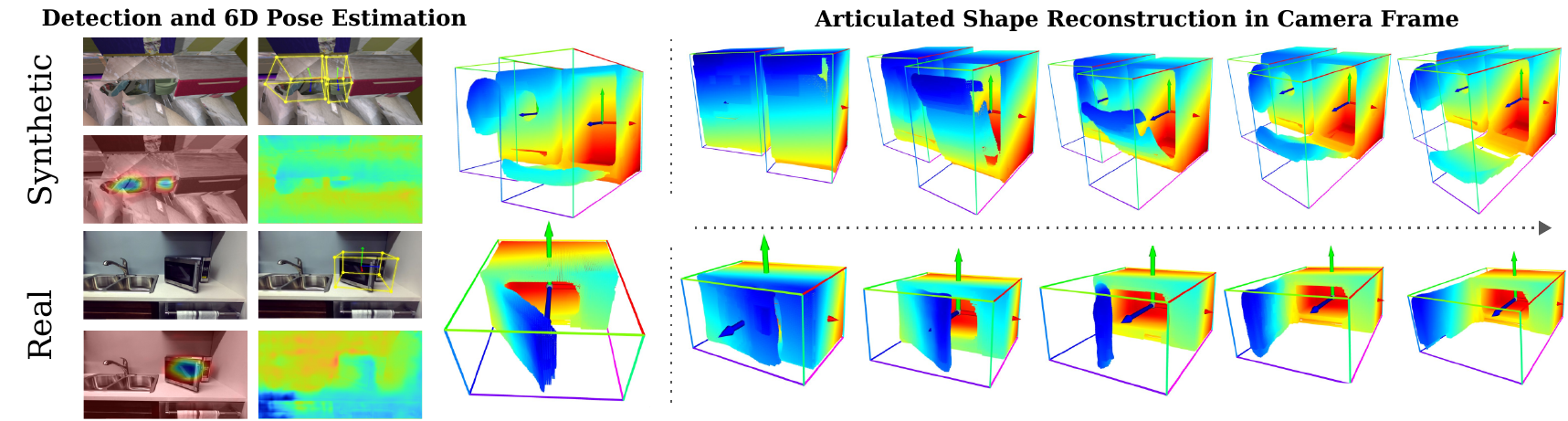}
    \captionsetup{width=\linewidth}
    \captionof{figure}{
    Visualization of \ourName{} on unseen object instances. We first use \ourName{} to jointly detect all objects in the scene and then articulate them while keeping the predicted shape code constant. 
    }
   \label{fig:teaser}
\end{center}%
}]


\begin{abstract}
   \input{content/0_abstract}
\end{abstract}

\input{content/1_introduction}
\input{content/2_related_work}

\input{content/3_method_inference}
\input{content/3_method_training}
\input{content/4_experiments}
\input{content/6_conclusion}

{\parskip=3pt
\noindent\textbf{Acknowledgements}: This work was partially funded by the Carl Zeiss Foundation with the ReScaLe project. 
}

{\small
\bibliographystyle{ieee_fullname}
\bibliography{FullArXivPaper}
}


\cleardoublepage

\begin{strip}
\begin{center}
\vspace{-5ex}
\textbf{\Large \bf
\ourName: \ourNameFull
} \\
\vspace{2ex}

\Large{\bf Supplementary Material}\\
\vspace{0.4cm}
\large{
Nick~Heppert,~
Muhammad~Zubair~Irshad,~
Sergey~Zakharov,~
Katherine~Liu,\\
Rares~Andrei~Ambrus,~
Jeannette~Bohg,~
Abhinav~Valada,~
Thomas~Kollar
}
\end{center}
\end{strip}

\appendix
\setcounter{section}{0}
\setcounter{equation}{0}
\setcounter{figure}{0}
\setcounter{table}{0}
\makeatletter
\renewcommand{\thesection}{S.\arabic{section}}
\renewcommand{\thesubsection}{S.\arabic{section}.\arabic{subsection}}
\renewcommand{\thetable}{S.\arabic{table}}
\renewcommand{\thefigure}{S.\arabic{figure}}
\renewcommand{\theequation}{S.\arabic{equation}}

\normalsize
\let\cleardoublepage\clearpage

\input{content/supp/0_abstract}
\input{content/supp/A_architectures}
\input{content/supp/B_backward_optim}
\input{content/supp/C_inverse_joint_decoder}
\input{content/supp/D_datasets}

\input{content/supp/E_additional_results}

\end{document}

%% file: content/assets/variable_definitions.tex
\newcommand{\ourName}{CARTO}
\newcommand{\ourNameFull}{Category and Joint Agnostic Reconstruction of ARTiculated Objects}

\newcommand{\latentVariable}{\boldsymbol{z}}
\newcommand{\latentShape}{\latentVariable_{\text{s}}}
\newcommand{\latentJoint}{\latentVariable_{\text{j}}}
\newcommand{\geometryDecoder}{\phi_{\text{geom}}}
\newcommand{\jointDecoder}{\phi_{\text{joint}}}

\newcommand{\jointType}{\textit{jt}}
\newcommand{\jointState}{q}

\newcommand{\objectCount}{M}
\newcommand{\objectIndex}{\MakeLowercase{\objectCount}}
\newcommand{\jointCount}{N}
\newcommand{\jointIndex}{\MakeLowercase{\jointCount}}

\newcommand{\object}{x}
\newcommand{\objectLoD}{l}

\newcommand{\spaceCoordinate}{\boldsymbol{x}}
\newcommand{\normal}{\boldsymbol{n}}
\newcommand{\sdfValue}{s}
\newcommand{\allSdfValues}{\boldsymbol{S}}

\newcommand{\realspace}{\mathbb{R}}
\newcommand{\dimensionsShapeSpace}{D_s}
\newcommand{\dimensionsJointSpace}{D_j}
\newcommand{\dimensionsIndex}{d}

\newcommand{\leakyClamp}{\text{clamp}_l}
\newcommand{\leakyThreshold}{\delta}
\newcommand{\leakySlope}{\alpha}

\newcommand{\lossGeneralFunction}{\mathcal{L}}
\newcommand{\lossScaling}{\delta}

\newcommand{\indexReconstruction}{\text{rec}}
\newcommand{\indexCodeRegularizer}{\text{reg}}
\newcommand{\indexJointRegularizer}{\text{jr}}
\newcommand{\indexJointType}{\jointType}
\newcommand{\indexJointState}{\jointState}

\newcommand{\similarityFunction}{\text{sim}}
\newcommand{\indexLatent}{\text{latent}}
\newcommand{\indexReal}{\text{real}}
\newcommand{\similarityFunctionLatent}{\similarityFunction_\indexLatent{}}
\newcommand{\similarityFunctionReal}{\similarityFunction_\indexReal{}}
\newcommand{\similarityMatrix}{\boldsymbol{S}}

\newcommand{\jointToCodeFunction}{\xi_{\text{code}}}
\newcommand{\polynomDegree}{p}

\newcommand{\importance}{\psi}

\newcommand{\depthMap}{D^{W \times H}}
\newcommand{\inputImage}{I^{W \times H \times 6}}

\newcommand{\sdfValueThreshold}{\epsilon}

\newcommand\+{\mkern2mu}

%% file: content/0_abstract.tex
We present \ourName{}, a novel approach for reconstructing multiple articulated objects from a single stereo RGB observation. We use implicit object-centric representations and learn a single geometry and articulation decoder for multiple object categories. Despite training on multiple categories, our decoder achieves a comparable reconstruction accuracy to methods that train bespoke decoders separately for each category. Combined with our stereo image encoder we infer the 3D shape, 6D pose, size, joint type, and the joint state of multiple unknown objects in a single forward pass. Our method achieves a 20.4\% absolute improvement in mAP 3D IOU50 for novel instances when compared to a two-stage pipeline. Inference time is fast and can run on a NVIDIA TITAN XP GPU at 1 HZ for eight or less objects present. While only trained on simulated data, \ourName{} transfers to real-world object instances. Code and evaluation data is available at: \href{http://carto.cs.uni-freiburg.de}{\url{carto.cs.uni-freiburg.de}}

%% file: content/1_introduction.tex
\section{Introduction}
\label{sec:intro}

Reconstructing 3D shapes and inferring the 6D pose and sizes of objects from partially observed input observations remains a fundamental problem in computer vision with applications in Robotics~\cite{mees2019self, jiang_ditto_2022, laskey2021simnet, jiang2021synergies} and AR/VR~\cite{zhang2021holistic, irshad_2022_shapo}. This object-centric 3D scene understanding problem is challenging and under-constrained since inferring 6D pose and shape can be ambiguous without prior knowledge about the object of interest.

Previous work has shown that it is possible to perform category-level 3D shape reconstruction and 6D pose estimation in real-time~\cite{irshad_centersnap_2022}, enabling the reconstruction of complete, fine-grained 3D shapes and textures. However, there are a wide variety of real-world objects that do not have a constant shape but can be articulated according to the object's underlying kinematics. There has been great progress in articulated object tracking~\cite{weng_captra_2021, sturm_probabilistic_2011, heppert_category-independent_2022, jain_screwnet_2021} and reconstruction~\cite{jiang_ditto_2022, noguchi_watch_nodate} from a sequence of observations. However, a sequence of observations is cumbersome since it often requires prior interaction with the environment.  In contrast, object reconstruction from a single stereo image through inferring latent information about an object a priori enables both grasping and manipulation of previously unknown articulated objects.  Additionally, estimates from a single image can also serve as a good initial guess for object tracking approaches~\cite{weng_captra_2021}.

Previous approaches to articulated object reconstruction from a single observation use a two-stage approach~\cite{liu_akb-48_2022} where first objects are detected using, e.g., Mask-RCNN \cite{he2017mask}. Then, based on the detection output, object properties, e.g. part-poses and NOCS maps~\cite{wang_normalized_2019}, are predicted and the object is reconstructed using backward optimization~\cite{mu_a-sdf_2021}. Such an approach is complex, error prone, does not scale across many categories, and does not run in real-time.

To mitigate the aforementioned challenges, building on ideas from~\cite{irshad_centersnap_2022} - a single-shot approach to output complete 3D information (3D shape, 6D pose, and sizes of multiple objects) on a per-pixel manner - we present ``\ourNameFull" (\ourName{}). 
First, extending \cite{mu_a-sdf_2021}, we train a robust category- and joint-agnostic 3D decoder~\cite{fan2017point, park_deepsdf_2019, mu_a-sdf_2021} by learning disentangled latent shape and joint codes. The shape code encodes the canonical shape of the object while the joint code captures the articulation state of the object consisting of the type of articulation (e.g., prismatic or revolute) and the amount of articulation (i.e. joint state). To disentangle both codes we impose structure among our learned joint codes by proposing a physically grounded regularization term. 
Second, in combination with our stereo RGB image encoder we can do inference in a \textit{single-shot manner} to detect the objects' spatial centers, 6D poses and sizes as well as shape and joint codes. The latter two can then be used as input to our category- and joint-agnostic decoder to directly reconstruct all detected objects.

To evaluate \ourName{}, we first evaluate the reconstruction and articulation state prediction quality of our category- and joint-agnostic decoder and compare against decoders trained on a single category. In an ablation study we show the necessity of our proposed joint code regularization over naively training the joint codes. 
We then quantitatively compare our full pipeline to a two-stage approach on synthetic data and show qualitative results on a new real-world dataset.

Our main contributions can be summarized as follows:
\begin{itemize}[noitemsep, topsep=0pt]
    \item An approach for learning a shape and joint decoder jointly in a category- and joint-agnostic manner.
    \item A single shot method, which in addition to predicting 3D shapes and 6D pose, also predicts the articulation amount and type (prismatic or revolute) for each object. 
    \item Large-scale synthetic and annotated real-world evaluation data for a set of articulated objects across 7 categories. 
    \item Training and evaluation code for our method.
\end{itemize}

%% file: content/2_related_work.tex
\section{Related Work}
\label{sec:related_work}
Related work to \ourName{} includes neural fields for reconstruction, implicit reconstructions of non-rigid objects and articulated object detection, pose estimation, and reconstruction.

\noindent\textbf{Neural Fields for Reconstruction}:
Neural fields, i.e. coordinate-based multi-layer perceptrons~\cite{xie2021neural}, have become a popular method for reconstruction in recent years. These methods encode continuous functions that model various scene properties, such as Signed Distance~\cite{park_deepsdf_2019}, radiance~\cite{mildenhall2021nerf}, and occupancy~\cite{mescheder2019occupancy}. Variations of these include hybrid discrete-continuous representations that employ an external data structure, i.e. a grid or an octree, to partition the implicit function~\cite{peng2020convolutional,zakharovroad}. The encoded shape can then be extracted via sphere tracing~\cite{liu2020dist} after querying the implicit function repeatedly. Recent advances in differential rendering have enabled learning of shapes, as well as other scene properties such as appearance, only from images and without the need for 3D supervision~\cite{mildenhall2021nerf}. Our approach falls into the paradigm of using neural fields for articulated object reconstruction and further learns a complete system for detection, pose estimation, and articulated shape reconstruction from a single observation.

\noindent\textbf{Implicit Reconstruction of Non-Rigid Objects}:
Going beyond static scenes with rigid objects,~\cite{bozic2020deepdeform} handle dynamic scenes while~\cite{su_-nerf_2021,palafox2022spams} focus on reconstructing humans by leveraging their strong shape and kinematic prior as well as the amount of readily available datasets. \cite{yang_lasr_2021} propose a general reconstruction framework to reconstruct any non-rigid entity (i.e. humans, animals, or objects) given only an RGB-video without requiring a category-specific shape template, while~\cite{lei_cadex_2022} focus on point cloud input data and split the prediction into a canonical shape and a deformation field. One downside of general reconstruction methods is that they do not leverage the rigidity and kinematic constraints of articulated objects. To explicitly use such priors,~\cite{noguchi_watch_nodate} propose a method that processes a multi-view sequence of a moving object and discovers its parts as well as their respective reconstruction and kinematic structure in an unsupervised way. Going one step further,~\cite{wei_self-supervised_nodate} learn a shape and appearance prior for each category which allows them to model accurate reconstructions of articulated objects with only 6 given views. Similarly,~\cite{tseng_cla-nerf_2022} propose a Neural Radiance Field (NeRF) based method that can reconstruct objects on a category level given some images of an object. Also leveraging a learned shape prior over objects of the same category,~\cite{mu_a-sdf_2021} reconstructs articulated objects using only a single observation by optimizing for a latent shape code; similarly to~\cite{tseng_cla-nerf_2022}, their method is only tested on revolute objects.

Our approach also represents objects through low-dimensional codes. However, by disentangling the shape from the articulation state in our code, our method becomes category- and joint-agnostic.  Other multi-category models, such as \cite{jiang_ditto_2022}, reconstruct objects given a point cloud in two different articulation states, which limits the approach to objects with a single joint; \cite{nie_structure_2022} uses many observations before and after articulation to reconstruct an object. Most similar to our disentangled representation, \cite{xu_unsupervised_2022} learns a latent space in which part-pairs are close if one part-pair can be transformed into another through a valid joint transformation.

\noindent\textbf{Articulated Object Detection, Pose Estimation and Reconstruction}: 
Work in articulated object detection and pose estimation typically first requires the detection of individual parts and their respective poses from a sequence of images demonstrating the articulation \cite{weng_captra_2021, sturm_probabilistic_2011, heppert_category-independent_2022, jain_screwnet_2021, jiang_ditto_2022} or from a single image~\cite{liu_akb-48_2022, li_category-level_2020, michel_pose_2015}.~\cite{li_category-level_2020} combines this part-level view of articulated objects with a holistic object-centric view as done for rigid objects \cite{wang_normalized_2019}. Similarly, our work predicts the poses for articulated objects in the scene in a single pass from a single stereo or RGB-D image, without the need to detect individual parts first. Most similar to us,~\cite{irshad_centersnap_2022,irshad_2022_shapo} also detect the pose, shape and scale of multiple objects from an RGB-D observation via a single-stage approach. Both methods perform category-agnostic detection of unseen object instances at test time, however, they are limited to rigid objects, with~\cite{irshad_centersnap_2022} using a point cloud decoder for shape reconstruction, while~\cite{irshad_2022_shapo} employs a latent shape and appearance prior. Our method also performs category-agnostic object detection and reconstruction, and we extend~\cite{irshad_centersnap_2022,irshad_2022_shapo} to handle articulated objects of multiple types in a single network forward pass, thus enabling fast and accurate articulate shape, pose and size estimation from a single stereo image.

%% file: content/3_method_inference.tex
\section{Technical Approach}
\label{sec:method}

\begin{figure*}
    \centering
    \includegraphics[width=0.9\textwidth]{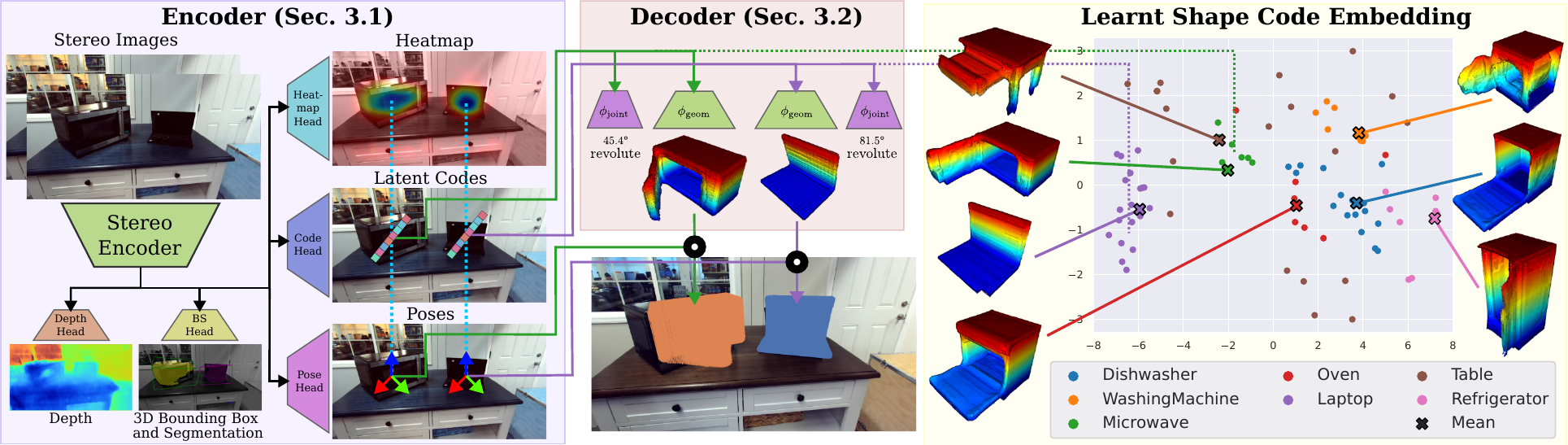}
    \caption{Overview of our proposed method. We first encode a stereo image using \cite{laskey2021simnet} to predict a depth and an importance value, a pose as well as a shape and joint code for each pixel using peak detection on the depth map allows us to detect objects which then can be reconstructed given the latent code. Last, to place the objects in camera frame we transform the reconstructed point cloud using the predicted poses at the peaks. On the right side we show the position of the predicted shape codes in a t-SNE visualization of the learned shape codes for the training used as input to our single category- and joint-agnostic decoder. We additionally project each categories mean shape code and use them to reconstruct the objects at the average prismatic and revolute joint state in the training set.}
    \label{fig:overview}
\end{figure*}

In this section, we detail our proposed single-shot detector for articulated objects. Our method consists of two individually learned components: an encoder that predicts latent object codes as well as poses in the camera frame and a decoder that reconstructs objects in a canonical frame that can be transformed into the camera frame through the predicted pose. An overview of our approach is shown in \cref{fig:overview}.

\subsection{Encoder}
\label{subsec:encoder}
Our encoder builds upon CenterSnap~\cite{irshad_centersnap_2022} and SimNet~\cite{laskey2021simnet}. For each pixel in our input stereo image $\inputImage$ we predict an importance scalar $\importance$, where a higher value indicates closeness to a 2D spatial center in the image of the object. The full output map of $\importance$ represents a heatmap over objects. 
Additionally, we predict a dense pixel map of canonical 6D poses for each articulated object independent of their articulation state. 
Further, we extend \cite{irshad_centersnap_2022}, which predicted a shape code $\latentShape \in \realspace^{\dimensionsShapeSpace}$, to also predict a joint code $\latentJoint \in \realspace^{\dimensionsJointSpace}$ for each pixel.  These codes can be used to predict the articulation state of the object.

Additionally, while not needed for our full pipeline, to guide the network towards important geometric object features, we also predict a semantic segmentation mask as well as 3D bounding boxes, again on a pixel-level. We use these predictions for constructing our baseline as described in \cref{subsubsec:full_pipeline_baseline}. Last, we also predict a depth map $\depthMap$. The full network architecture is given in \ifthenelse{\boolean{reftosupp}}{\cref{supp:subsec:encoder_architecture}}{Sec.~S.1.1}. 

During inference of our full pipeline, given the predicted heatmap of importance values, we use non-maximum suppression to extract peaks in the image. At each peak, we then query the feature map to get the pose, shape, and joint code. We convert our 13-dimensional pose vector to a scale value $\in\realspace$ of the canonical object frame, a position $\in\realspace^{3}$ and using \cite{bregier2021deepregression} to an orientation $\in\realspace^{3 \times 3}$ in the camera frame.
We then use the shape and joint code to reconstruct each object in its canonical object frame using our decoder. After reconstruction, we use the predicted pose to place the object in the camera frame as shown in \cref{fig:overview}. 

\subsection{Decoder}
\label{subsec:decoder}
Given a latent code, the decoder reconstructs object geometry, classifies the discrete joint type (i.e., as prismatic or revolute), and predicts the continuous joint state.
To disentangle the shape of the object from its articulation state, we split the latent code in two separate codes: a shape and a joint code. We assign the same unique shape code  $\latentShape \in \realspace^{\dimensionsShapeSpace} $ to an object instance in different articulation states, where an articulation state is expressed through its own joint code variable $\latentJoint \in \realspace^{\dimensionsJointSpace}$. We structure our decoder as two sub-decoders, one for reconstructing the geometry (\cref{subsubsec:geometry_decoder}) and the other for predicting the joint type $\jointType$ and state $\jointState$ (\cref{subsubsec:joint_decoder}). See \ifthenelse{\boolean{reftosupp}}{\cref{supp:subsec:decoder_architecture}}{Sec.~S.1.2} for a full architecture description.

\subsubsection{Geometry Decoder}
\label{subsubsec:geometry_decoder}
The geometry decoder $\geometryDecoder$ reconstructs objects based on a shape code $\latentShape$ and joint code $\latentJoint$. In principle, the approach is agnostic to the specific decoder architecture as long as it is differentiable with respect to the input latent codes. While there are many potential options such as occupancy maps \cite{mescheder2019occupancy} as adopted in \cite{jiang_ditto_2022}, we use signed distance functions (SDFs) \cite{park_deepsdf_2019} due to the proven performance in \cite{mu_a-sdf_2021, park_deepsdf_2019}.
Specifically, in the case when using SDFs as our geometry decoder, the model takes as input a point in 3D space $\spaceCoordinate$ as well as a shape  $\latentShape$ and joint code $\latentJoint$
\begin{equation}
    \geometryDecoder(\latentShape, \latentJoint, \spaceCoordinate) = \hat{\sdfValue}_{\spaceCoordinate}
\end{equation}
and predicts a value $\hat{\sdfValue}_{\spaceCoordinate}$ that indicates the distance to the surface of the object.

For faster inference, we implement a multi-level refinement procedure~\cite{irshad_2022_shapo}. We first sample query points on a coarse grid and refine them around points that have a predicted distance within half of the boundary to the next point. This step can be repeated multiple times to refine the object prediction up to a level $\objectLoD$. Eventually, we extract the surface of the objects by selecting all query points $\spaceCoordinate$ for which $ |\hat{\sdfValue}_{\spaceCoordinate}| < \sdfValueThreshold $ holds. By taking the derivative 
\begin{equation}
    \normal_{\spaceCoordinate} = \frac{\partial \geometryDecoder(\latentShape, \latentJoint, \spaceCoordinate)}{\partial \spaceCoordinate}
\end{equation}
and normalizing it we get the normal $\hat{\normal}_{\spaceCoordinate}$ at each point $\spaceCoordinate$, which can then be used to project the points onto the surface of the object with $\hat{\spaceCoordinate} = \spaceCoordinate - \hat{\sdfValue}_{\spaceCoordinate} \hat{\normal}_{\spaceCoordinate}$.

\subsubsection{Joint Decoder}
\label{subsubsec:joint_decoder}
As we represent the articulation state of the object implicitly through a joint code $\latentJoint$, we additionally introduce an articulation state decoder $\jointDecoder$ to regress a discrete joint type $\jointType = \{ \text{prismatic}, \text{revolute} \}$ and a continuous joint state $q$:
\begin{equation}
    \jointDecoder(\latentJoint) = \hat{\jointType}, \hat{\jointState}
\end{equation} 
We use a multi-layer perceptron with 64 neurons in one hidden layer. 

\begin{figure}
    \centering
    \includegraphics[width=0.7\linewidth]{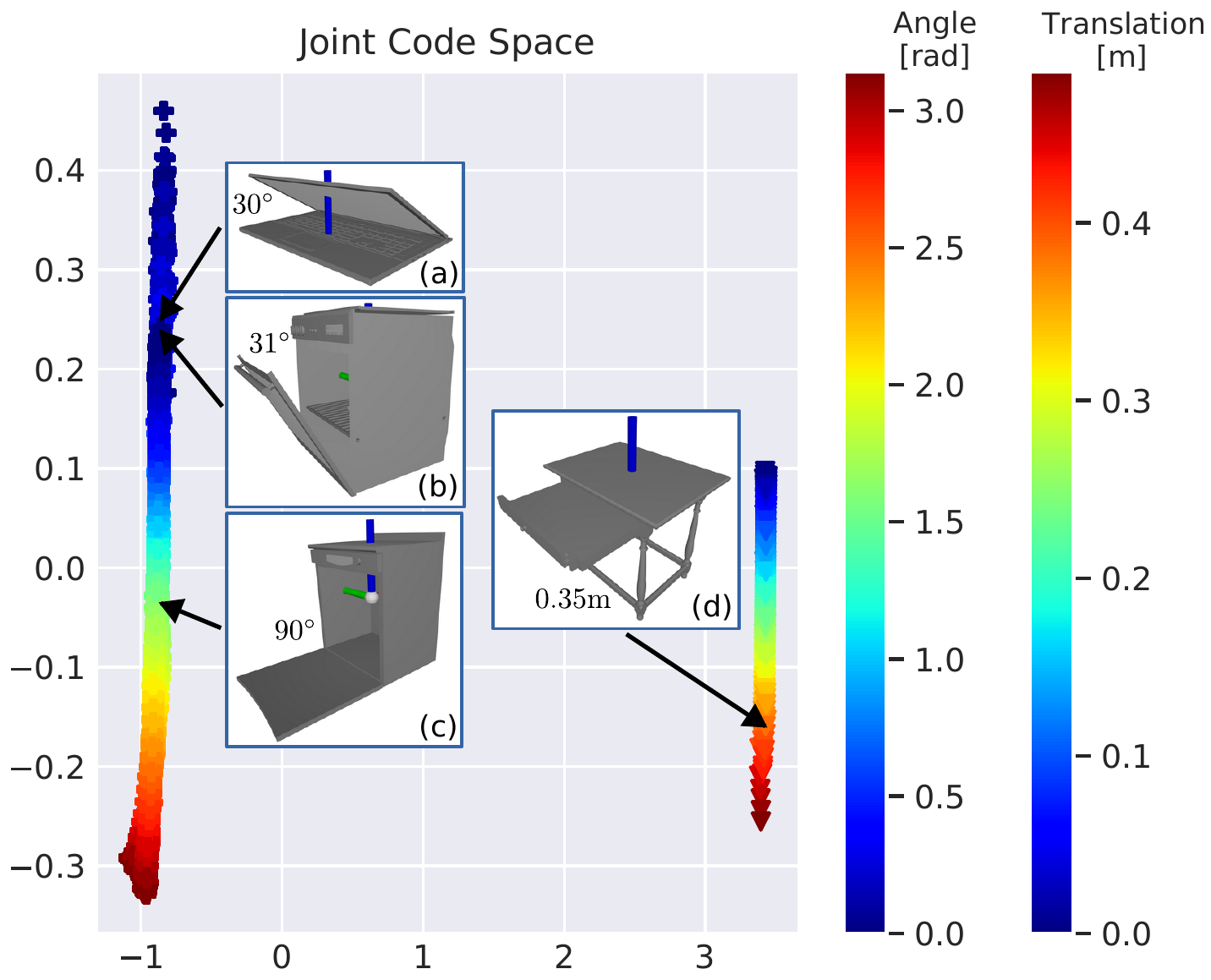}
    \caption{Intuition for Latent Space Regularization.
    Our main idea is that the joint codes 
    of two similarly articulated objects should be close. We define the similarity first through the joint type $\jointType$ and second through an exponential distance measure of the joint state $\jointState$. Here, the laptop (a) and the oven (b) have a revolute joint and are similarly wide open around $30\degree$. Compared to that, table (d) has a prismatic joint and thus should not be close to the revolute instances. Contrary to that, the dishwasher (c), has a revolute joint but is opened much more than the other revolute objects and therefore, should be relatively close. The visualization shows a lower dimensional projection of our learned latent joint space trained using our regularization.}
    \label{fig:latent_space_intuition}
\end{figure}

\subsubsection{Backward Code Optimization}
\label{para:backward_code_optimiziation}
For the task of canonical object reconstruction from a set of SDF values at specific query points, we follow the optimization procedure from \cite{mu_a-sdf_2021} to retrieve a shape and joint code. Different from \cite{mu_a-sdf_2021}, we utilize GPU parallelization to optimize multiple code hypotheses at once and pick the best one in the end. Additionally, we do not reset the codes as done in \cite{mu_a-sdf_2021} but rather freeze them for some iterations. In early testing, we discovered that first optimizing for both codes jointly together gives a good initial guess. Freezing the joint code in a second stage of optimization helps such that the shape fits the static part and then last, freezing the shape code such that the joint code can do some fine adjustment to the articulation state of the object. To guide the gradient in the joint code space, we first transform the space using the singular value decomposition of the stacked training joint codes $\latentJoint^{\jointIndex} \+\forall\+ \jointIndex \in 1, \ldots, \jointCount$.
For further information, refer to \ifthenelse{\boolean{reftosupp}}{\cref{supp:sec:backward_optim}}{Sec.~S.2}. 

%% file: content/3_method_training.tex
\subsection{Training Protocol}
To train \ourName{}, we first train the decoder as it provides ground truth training labels for the shape and joint code supervision of our encoder. Once shape and joint code labels are obtained for the objects in the dataset, we then train our encoder to predict the latent codes in addition to object pose. Thus, to adhere with our training procedure, we will first explain how to train our decoder and then our encoder. 
\subsubsection{Decoder}
Given a training set of $\objectCount$ objects each in $\jointCount$ articulation states, we denote the $\objectIndex$-th object in its $\jointIndex$-th articulation state as $\object_{\objectIndex, \jointIndex}$. As during training, a fixed association between each object and its latent codes is given, we can uniquely identify $\object_{\objectIndex, \jointIndex} = ( \latentShape^{\objectIndex}, \latentJoint^{\jointIndex})$ as a tuple of both. This allows us to pass the gradient all the way to the codes themselves and thus, the embedding spaces rearrange them accordingly. Similar to \cite{mu_a-sdf_2021}, during training we regularize the codes through minimizing the L2-norm
\begin{equation}
    \label{eqn:loss_code_reg}
    \lossGeneralFunction_{\indexCodeRegularizer}(\latentVariable) =
    \| \latentVariable \|,
\end{equation}
where $\latentVariable$ is either $\latentShape$ or $\latentJoint$.

{\parskip=5pt
\noindent\textbf{The geometry decoder} described in \cref{subsubsec:geometry_decoder} is trained on a set of query points $\spaceCoordinate$ close to the object surface sampled as in \cite{park_deepsdf_2019}.}
We define our reconstruction loss $\lossGeneralFunction_{\indexReconstruction}$ as in \cite{park_deepsdf_2019} but use a leaky clamping function \begin{equation}
    \leakyClamp(s|\leakyThreshold, \leakySlope) = 
    \begin{cases} 
        s & |s| \leq \leakyThreshold, \\
        \leakySlope s &  |s| > \leakyThreshold
    \end{cases}
\end{equation}
which is conceptually similar to a leaky ReLU by instead of hard clamping values above a threshold $\leakyThreshold$, we multiply it by a small factor $\leakySlope$. Initial testing revealed a more stable training. 
Our reconstruction loss at one query point $\spaceCoordinate$ is now given by:
\begin{align}
    \label{eqn:loss_reconstruction}
    & \lossGeneralFunction_{\indexReconstruction}\left( \latentShape, \latentJoint, \spaceCoordinate, \sdfValue_{\spaceCoordinate} \right) = \\
    & \quad \left| 
        \leakyClamp(
            \geometryDecoder(\latentShape, \latentJoint, \spaceCoordinate)
            | \leakyThreshold, \leakySlope
        )
        -
        \leakyClamp(
            \sdfValue_{\spaceCoordinate}
            | \leakyThreshold, \leakySlope
        )
    \right|, \nonumber
\end{align}
where $\sdfValue_{\spaceCoordinate}$ is the ground truth distance to the surface.

{\parskip=5pt
\noindent\textbf{The joint decoder} introduced in \cref{subsubsec:joint_decoder} is jointly trained with the aforementioned geometry decoder. For the joint type loss $\lossGeneralFunction_{\indexJointType}$, we use cross entropy between the predicted joint type $\hat{\jointType}$ and ground truth $\jointType$ and for the joint state loss $\lossGeneralFunction_{\indexJointState}$ the L2-norm between the predicted joint state $\hat{\jointState}$ and ground truth $\jointState$.}

{\parskip=5pt
\noindent\textbf{Joint Space Regularization}: 
\label{subsubsec:joint_space}
One core contribution of our approach is how we impose structure in our joint code space during our decoder training. Here, we enforce the same similarity of latent codes as their corresponding articulations states have. A visualization of the underlying idea is shown in \cref{fig:latent_space_intuition}.
Formally, given the joint codes $\latentJoint^{k}$ and $\latentJoint^{l}$ encoding two different articulation states $k, l \in 1, \ldots, \jointCount$, we define the similarity between them in latent space as 
\begin{equation}
    \similarityFunctionLatent \left( \latentJoint^{k}, \latentJoint^{l} \right) =
    \exp \left(- \frac{\| \latentJoint^{k} - \latentJoint^{l} \| }{\sigma} \right). 
\end{equation}
Similarly, we define the respective similarity in real joint space, considering the joint types $\jointType^k$ and $\jointType^l$ and the joint states $\jointState^k$ and $\jointState^l$, through
\begin{align}
    \label{eqn:joint_sim_real}
    &\similarityFunctionReal \left( \left( \jointType^k, \jointState^k \right), \left( \jointType^l, \jointState^l \right) \right) = \\
    &\qquad\qquad \begin{cases}
        \exp \left(
            - \left( \frac{\jointState^k-\jointState^l}{\sigma_\jointType} \right)^2 
        \right) & \jointType^k = \jointType^l \\
        0 & \jointType^k \neq \jointType^l, \nonumber
    \end{cases}
\end{align}
where $\sigma_\jointType$ is a joint type specific scaling.
By minimizing the L1-norm between both similarity measurements 
\begin{align}
    \label{eqn:joint_code_reg_single}
    & \lossGeneralFunction_{\indexJointRegularizer} \left( 
        \latentJoint^{k}, \latentJoint^{l} \right) =   \\
    & \quad \left|
            \similarityFunctionLatent \left( \latentJoint^{k}, \latentJoint^{l} \right)
            -
            \similarityFunctionReal \left( \left( \jointType^k, \jointState^k \right), \left( \jointType^l, \jointState^l \right) \right)
        \right|  \nonumber
\end{align}
we enforce that the latent similarities are similarly scaled as their real similarities. 

We scale this formulation to all articulation states in the training set as described below. Calculating $\similarityFunctionReal$ can be done once in a pre-processing step for all articulation state pairs $k,l \in 1, \ldots, \jointCount$ resulting in a matrix $\similarityMatrix_\indexReal \in \realspace^{\jointCount \times \jointCount}$. Similarly, calculating all $\similarityFunctionLatent$-pairs can be efficiently implemented as a vector-vector product. We denote the resulting matrix as $\similarityMatrix_\indexLatent \in \realspace^{\jointCount \times \jointCount}$. \cref{eqn:joint_code_reg_single} now simplifies to\looseness=-1
\begin{align}
    \label{eqn:joint_code_reg}
    \lossGeneralFunction_{\indexJointRegularizer} =  
    \frac{
    \left|
        \similarityMatrix_\indexLatent
            -
        \similarityMatrix_\indexReal
    \right|}{\jointCount^2}.
\end{align}
Through this efficient calculation, optimizing this loss term comes with almost no overhead during training. This concept of similarity can be extended for arbitrary kinematic graphs.
}

{\parskip=5pt
\noindent\textbf{Pre-Training}: Before we start training our full decoder, we only optimize our joint codes $\latentJoint^{\jointIndex}\+\forall\+\jointIndex \in 1,\ldots, \jointCount$. The pre-training helps with learning the full decoder as our joint codes are already more structured and thus, it is easier to learn the shape and joint code disentanglement. 
In the pre-training, we minimize
\begin{equation}
    \lossGeneralFunction_{\text{pre}} = 
    \lossScaling_{\indexCodeRegularizer, \latentJoint, \text{pre}} \lossGeneralFunction_{\indexCodeRegularizer, \latentJoint}
    + \lossScaling_{\indexJointRegularizer, \text{pre}} \lossGeneralFunction_{\indexJointRegularizer},
\end{equation}
where $\lossGeneralFunction_{\indexCodeRegularizer, \latentJoint}$ is the default norm regularization from \cref{eqn:loss_code_reg} and $\lossGeneralFunction_{\indexJointRegularizer}$ was introduced in \cref{eqn:joint_code_reg}.
}

{\parskip=5pt
\noindent\textbf{Loss Function}: Given an object $\object_{\objectIndex, \jointIndex}$, we express our full decoder loss as} 
\begin{align}
    \label{eqn:full_loss}
    \lossGeneralFunction = &
    \lossScaling_{\indexCodeRegularizer, \latentShape} \lossGeneralFunction_{\indexCodeRegularizer, \latentShape}
    + \lossScaling_{\indexCodeRegularizer, \latentJoint} \lossGeneralFunction_{\indexCodeRegularizer, \latentJoint}
    + \lossScaling_{\indexReconstruction} \lossGeneralFunction_{\indexReconstruction} \\
    & \quad+ \lossScaling_{\indexJointType} \lossGeneralFunction_{\indexJointType}
    + \lossScaling_{\indexJointState} \lossGeneralFunction_{\indexJointState} , \nonumber
\end{align}
where $\lossGeneralFunction_{\indexCodeRegularizer, \latentVariable}$ are the shape and joint code regularization from \cref{eqn:loss_code_reg}, $\lossGeneralFunction_{\indexReconstruction}$ is the reconstruction loss introduced in \cref{eqn:loss_reconstruction}, $\lossGeneralFunction_{\indexJointType}$ and $\lossGeneralFunction_{\indexJointState}$ are the joint type and state loss.
We jointly optimize $\lossGeneralFunction$ for the latent shape and joint code as well as the network parameters of the geometry decoder and joint decoder
using ADAM \cite{Kingma2015AdamAM} for 5000 epochs.
Our new joint code regularizer loss $\lossGeneralFunction_{\indexJointRegularizer}$ introduced in \cref{eqn:joint_code_reg} is minimized at the end of each epoch separately scaled by $\lossScaling_{\indexJointRegularizer}$. All $\lossScaling$ variables are scalars to balance the different loss terms and are reported in \ifthenelse{\boolean{reftosupp}}{\cref{supp:tab:decoder_scaling_training}}{Tab.~S.1}.

\subsubsection{Encoder}
Using \cite{laskey2021simnet} we generate a large-scale dataset in which we annotate each pixel with its respective ground truth value from the simulation as described in \cref{subsec:encoder}. For annotating the shape codes we directly use the results of our previous encoder training whereas for the joint code we use 
our inverse mapping explained in \cref{subsec:inverse_joint_decoder} to retrieve joint codes for arbitrary sampled articulation states.

\subsubsection{Inverse Joint Decoder}
\label{subsec:inverse_joint_decoder}
To solve the inverse problem, given an articulation state for which we want to retrieve a joint code, we fit polynomial functions in the learnt joint code space. With the help of this mapping, we can retrieve arbitrary joint codes which then can be combined with a shape code to reconstruct objects in novel articulation states which have not been seen during the decoder training. Additionally, the mapping provides joint code training labels for the encoder. 

We describe the full mapping as a function $\jointToCodeFunction (\jointType, \jointState) = \latentJoint$ that takes a joint type $\jointType$ and joint state $\jointState$ as input and outputs a joint code $\latentJoint$. We leverage the fact that after decoder training, we learned a joint code $\latentJoint^{\jointIndex}$ for each known training articulation state. We now define individual mappings for each joint type $\jointType$ the following way. We will treat each latent dimension $\dimensionsIndex$ separately. For each dimension $\dimensionsIndex$, we fit a polynomial function $\jointToCodeFunction^{\jointType, \dimensionsIndex} (\jointState)$ of varying degree $\polynomDegree$ through all point tuples $(\jointState^{\jointIndex}, \latentJoint^{\jointIndex}(\dimensionsIndex)) \+\forall\+ \jointIndex \in 1, \ldots, \jointCount$. 

The final function
\begin{equation}
    \jointToCodeFunction (\jointType, \jointState) = 
    \begin{bmatrix}
        \jointToCodeFunction^{\jointType, 1} (\jointState) \\
        \vdots \\
        \jointToCodeFunction^{\jointType, \dimensionsJointSpace} (\jointState)
    \end{bmatrix}
\end{equation}
is then given by evaluating the polynomials individually and stacking the results into a vector.
The exact choice of $\polynomDegree$ is not important as long as the amount of joint codes to fit to is much higher than the potential dimensions of the polynomial $\polynomDegree \ll \jointCount$. Thus, we fixed $\polynomDegree = 5$ for all of our experiments. 
A visualization of our learned latent joint space and the fitted polynomials is given in \ifthenelse{\boolean{reftosupp}}{\cref{supp:subfig:carto_latent_joint_space}}{Fig.~S.3a}.

%% file: content/4_experiments.tex
\section{Experiments}
\label{sec:experiments}
We conduct two main experiments, an object-centric canonical reconstruction task and a full scene reconstruction task. The first experiment is to evaluate the performance of our newly introduced decoder while the second experiment highlights the advantages of our single-forward pass method compared to a two-stage approach.

\subsection{Object Set}
\label{subsec:object_set}
For both experiments, we use 3D models from the PartNet-Mobility \cite{Xiang_2020_SAPIEN} object set to generate a training and test data set.

{\parskip=5pt
\noindent\textbf{Categories}:
While PartNet-Mobility provides more than 2000 objects from 46 categories, as done in previous work~\cite{mu_a-sdf_2021, heppert_category-independent_2022, jain_screwnet_2021, xue_omad_2021, li_category-level_2020} we only select a subset of all categories. From this subset of categories, we select objects with one fixed base part and one significant moving part (e.g. we filter out knobs, buttons etc.). To later create realistic room layouts, we further differentiate between three placement types for objects, stand-alone (SA), counter (C) and table-top (TT) objects. In \cref{tab:object_set} we list the number of objects per category we selected as well as the context in which they can be used.}

\begin{table}
    \centering
    \small
    \caption{Overview of Selected Objects. We select a subset of the PartNet-Mobility \cite{Xiang_2020_SAPIEN} object set and report the amount of instances we selected per category for our training and test set. Our final object set has 92 objects instances for training and 25 for testing.}
    \label{tab:object_set}
    \begin{tabular}{llccccc}
        \toprule
        Category & \makecell{Joint\\Type} & SA & C & TT & Train & Test \\
        \midrule
        Dishwasher & Rev. & \xmark & \cmark & \xmark & 18 & 5 \\
        Laptop & Rev. & \xmark & \xmark & \cmark & 20 & 5 \\
        Microwave & Rev. & \cmark & \xmark & \cmark & 10 & 5 \\
        Oven & Rev. & \cmark & \cmark & \xmark & 7 & 3 \\
        Refrigerator & Rev. & \cmark & \cmark & \xmark & 10 & 2 \\
        Table & Pris.& \cmark  & \xmark & \xmark & 19 & 5 \\
        WashingMachine & Rev. & \cmark & \cmark & \xmark & 8 & 2 \\
        \bottomrule
    \end{tabular}
    \vspace{0.1cm}
    \\{\textit{SA = stand-alone, C = counter, TT = table-top\\Rev. = Revolute, Pris. = Prismatic}}
    \vspace{-0.3cm}
\end{table}

{\parskip=5pt
\noindent\textbf{Object Canonicalization}: \label{subsubsec:object_canonicalization}
When tackling the task of reconstructing objects in a canonical object frame, usually, objects are canonicalized such that they either fit into the unit cube or unit sphere. This helps with the stability of learning and simplifies hyperparameter tuning. This approach fails for articulated objects as their outer dimensions change depending on the joint state. Blindly rescaling an articulated object such that it fits inside the unit cube or unit sphere results in an inconsistent part scaling across different joint states. To mitigate this problem, \cite{li_category-level_2020} proposed the NAOCS-space. Following their approach, first, we bring an object in its closed state (i.e. the lower joint limit) and put it in a canonical orientation such that for all objects, Z is pointing upwards and X back, Y is given through a right-hand coordinate frame system. Different from\cite{li_category-level_2020}, we rescale and translate the closed object such that it fits in the unit cube and then backwards apply that same rescaling and translation to all objects of the same instance independent of the joint state of the object.
It is important to note that rescaling an articulated object has no impact on revolute joint states (in~$\deg$), but prismatic joint states (in~$\si{m}$) which have to be rescaled accordingly.
}
\subsection{Canonical Reconstruction Task}
\label{subsec:canonical_reconstruction_task}
In our first experiment, we evaluate how well our decoders reconstruct the object's geometry and the articulation state. Thus, the task is not to reconstruct the object in the camera frame, but simply in its canonical frame. As described in \cref{para:backward_code_optimiziation}, we will optimize the shape and joint code for each input SDF with ADAM \cite{Kingma2015AdamAM} first jointly for 400 steps, then only the shape code for 100 steps and finally, only the joint code for 100 steps.

{\parskip=5pt
\noindent\textbf{Dataset}:
To generate our dataset for the canonical reconstruction task, we first apply the aforementioned canonicalization to each object from our object set described in \cref{subsec:object_set}. Here, the placement type does not matter. Then, we sample each object in 50 joint configurations uniformly spaced within the joint limits, make the meshes watertight using \cite{DBLP:journals/corr/abs-1802-01698} and follow \cite{park_deepsdf_2019} to generate 100k ground truth SDF values. Lastly, we rescale the generated data by the largest extents across all meshes to a unit cube. As mentioned in \cref{subsubsec:object_canonicalization} we also have to rescale prismatic joint states accordingly.
While we do not consider this as a new dataset contribution, we make our generation code available to allow other researchers to adjust the selected object instances and categories and generate their own data.

{\parskip=5pt
\noindent\textbf{Baselines and Ablations}:
\label{subsubsect:canonical_reconstruction_baseline_ablations}
Throughout this experiment, we will compare against the state-of-the-art for category-level object reconstruction method \textit{A-SDF} \cite{mu_a-sdf_2021}. As \textit{A-SDF} is designed to work on a single category, first, to show that learning an implicit joint code does not have a negative impact rather than using the real joint state directly as input, we compare against \textit{A-SDF} directly by training \textit{\ourName{}} also only on a single category. 
Second, we will jointly train \textbf{\textit{}} on all categories to highlight that \textit{\ourName{}} is able to generalize to a wide variety of categories and articulations using one model.
Third, we additionally perform an ablation study to understand the importance of our similarity regularization introduced in \cref{subsubsec:joint_space}. In this ablation study, we remove the pre-training step and the post-epoch step. We call this model \textit{\ourName-No-Enf}. And fourth, we extend \textit{A-SDF} to also take the joint type as input which allows us to train it jointly on all categories.

\noindent Please note that we neglect \textit{A-SDF}s proposed test-time adaption (TTA) technique as in real applications it would not be practical to keep network weights of all different instances encountered. Results using TTA are reported in \ifthenelse{\boolean{reftosupp}}{\cref{supp:subsec:a-sdf_tta}}{Sec.~S.5.1}. 
}

{\parskip=5pt
\noindent\textbf{Metrics}:
To measure reconstruction quality, we report the bi-directional L2-Chamfer distance multiplied by 1000 between the ground truth points and the extracted points using the model's respective generation method. To quantify the articulation state prediction, we will report the joint type prediction accuracy as well as the joint state error measured in~$\deg$~or~$\si{m}$~depending on the joint type for all correctly classified joint types.}

\input{content/assets/tables/canonical_reconstruction_small}

{\parskip=5pt
\noindent\textbf{Results}: The results for the canonical reconstruction task are shown in \cref{tab:decoder_results_short}. While no method clearly outperforms the other, it is notable though that \ourName{} and \textit{A-SDF} trained on all categories are performing slightly better on average across all categories compared to our baselines. This shows that having disentangled joint and shape codes in \ourName{}
can make the reconstruction category-agnostic.}
\subsection{Full Pipeline Task}
\label{subsec:full_pipeline_task}
In our second experiment, the full pipeline task, we want to investigate the advantages \ourName{} has over a two-stage approach. To that end, we set up two experiments, one on simulated data and one on real-world data. For the full pipeline experiment, we use the trained decoders from the previous experiment.

{\parskip=5pt
\noindent\textbf{Datasets}:
We evaluate on two datasets, quantitatively on a synthetic dataset that aligns with our synthetic training dataset and qualitatively on a newly collected real-world dataset.}

{\parskip=0pt
\textit{Synthetic Data}: For training our encoder, we use SimNet \cite{laskey2021simnet} to generate a large-scale dataset of indoor kitchen environments. We use the same articulated object instances from \cref{tab:object_set} we also used to train our decoders. Unlike the previous experiment, the placement type of the object matters here. For each randomly sampled articulated object in the scene, we randomly sample a joint state in its joint limits as well as a scale from a pre-defined scale for each category. 
To get ground truth joint codes for sampled articulation states we use the proposed method in \cref{subsec:inverse_joint_decoder}. After sampling a scene, we generate noisy stereo images as well as non-noisy depth images (only used for evaluation of the baseline).
To generate our synthetic test dataset we follow the same procedure with the only exception that we use the defined test-instances.}

{\parskip=0pt
\textit{Real Data}: Additionally, we evaluate the performance on a real-world test dataset we collected. Therefore, we select two real object instances from each of the following categories: knives, laptops, refrigerators, staplers, ovens, dishwashers, microwaves as well as one storage furniture and washing machine instance. We place these instances in common household environments. For each object, four different viewpoints were collected for each of the four articulation states for the object.  We measure the real joint state and annotate the data using~\cite{Sager_2022} with orientated 3D bounding boxes. In total, we collected 263 images. For collection we used a ZED 2 stereo camera. To get depth images we use state-of-the-art, off-the-shelf learned stereo depth methods to produce highly accurate depth images \cite{shankar2022learned}.
}

{\parskip=0pt
\textit{Comparison to other Datasets}: To the best of our knowledge, the closest works to \ourName{}'s dataset are the RBO~\cite{martin-martin_rbo_2018} and BMVC~\cite{michel_pose_2015} dataset. Both datasets do not provide large-scale synthetic stereo-RGB or RGB-D images and only half of the categories with readily available 3D furniture models from PartNetMobility. For a full comparison see \ifthenelse{\boolean{reftosupp}}{\cref{tab:dataset_comparison}}{Tab.~S.2}.
}

{\parskip=5pt
\noindent\textbf{Baselines}:
\label{subsubsec:full_pipeline_baseline}
We set up two baselines using A-SDF \cite{mu_a-sdf_2021} and follow their proposed method to reconstruct objects in the camera frame. Since A-SDF assumes a given segmentation of the object as well as a pose that transforms the object from the camera frame to the object-centric frame, we will compare against two versions of A-SDF. One, where we use ground truth segmentation masks and poses which we call \textit{A-SDF-GT} and one, where we use our model to predict segmentation masks whose center we then use to query our pixel-level pose map. We call this variant simply \textit{A-SDF}.
In both cases we approximate normals using the depth image, use the segmentation masks to extract the corresponding object point cloud from the depth image, transform the point clouds into the canonical object frame using the predicted pose, create SDF values, optimize for the SDF values as done in \cref{subsubsect:canonical_reconstruction_baseline_ablations} and eventually reproject the reconstruction into the camera frame using the same transformation.
}

{\parskip=5pt
\noindent\textbf{Metrics}:
We compare our reconstructions in the camera frame using two different metrics typically used for object pose prediction \cite{wang_normalized_2019}. First, we compare the absolute error of the position and orientation by reporting the respective percentage below $10\degree10\si{cm}$ and $20\degree30\si{cm}$ combined. Second, we evaluate the average precision for various IOU-overlap thresholds~(\textbf{IOU25} and \textbf{IOU50}) between the reconstructed bounding box and the ground truth bounding box. Both metrics serve as a proxy for articulation state and reconstruction quality. For evaluation results on these more fine-grained metrics, we refer to \ifthenelse{\boolean{reftosupp}}{\cref{supp:subsec:extend_metrics}}{Sec.~S.5.1}. 
}

{\parskip=5pt
\noindent\textbf{Results}:
In \cref{tab:full_pipeline} we report results using the aforementioned metrics as well as show a speed comparison of \ourName{} against the baselines.
}
\begin{table}
    \centering
    \caption{Full Scene Reconstructions Results.}
    \label{tab:full_pipeline}
    \begin{subtable}[h]{0.45\textwidth}
        \footnotesize
        \centering
        \begin{tabular}{l|cccc}
        \toprule
            Method & IOU25 $\uparrow$ & IOU50 $\uparrow$ & $10\si{\degree}10\si{cm} \uparrow$ & $20\si{\degree}30\si{cm} \uparrow$  \\
            \midrule
            A-SDF-GT & 45.2 & 27.1 & N/A & N/A \\
            A-SDF & 33.9 & 10.4 & 27.1 & 70.8 \\
            \ourName & \textbf{64.0} & \textbf{31.5} & \textbf{28.7} & \textbf{76.6}\\
        \bottomrule
        \end{tabular}
        \caption{\footnotesize mAP Reconstruction Results.}
        \label{tab:full_pipeline_synthetic}
    \end{subtable}%
    \hfill
    \newline
    \vspace*{0.2cm}
    \newline
    \begin{subtable}[h]{0.45\textwidth}
        \footnotesize
        \centering
        \begin{tabular}{c|c|ccc|c}
            \toprule
            Method & Sample Grid & Det. & Optim. & Recon. & Total \\
            \midrule
            A-SDF & 256 & 5.390 & 21.600 & 7.836 & 64.262 \\
            \ourName & 256 & \textbf{0.264} & N/A & 0.414 & 1.092 \\
            \ourName & 128 & \textbf{0.264} & N/A  & \textbf{0.097} & \textbf{0.458} \\
            \bottomrule
        \end{tabular}
        \caption{\footnotesize Detection Speed of Approaches in [s]. We measure the speed of our approaches on a common desktop using a Nvidia Titan XP GPU. Sample grid defines how many points are sampled along each dimension. Total time assumes two detected objects. \textit{Det. = Detection time per image, Optim. = Optimiziation time per object, Recon. = Reconstruction time per object}}
        \label{tab:detection_speed}
    \end{subtable}
    \vspace{-0.4cm}
\end{table}

{\parskip=1pt
\noindent
\textit{Reconstruction}:
As visible in \cref{tab:full_pipeline_synthetic} \textit{\ourName{}} shows superior performance over both variants of \textit{A-SDF} for our full reconstruction task.
Overall the performance of all methods is lower compared to similar experiments on category-level rigid object detection \cite{wang_normalized_2019}. This can be attributed to the fact that in our kitchen scenarios we deal with heavy occlusions due to many objects being placed under the counter. Taking the occlusions into consideration, it becomes clear that for \textit{A-SDF} it is very difficult to estimate the exact extents given only a front-showing partial point cloud of the object. Compared to that, \textit{\ourName{}} benefits from its single-shot encoding step as the whole image is taken into consideration.
We show qualitative results on our synthetic and real-world data in \cref{fig:teaser} as well as in \ifthenelse{\boolean{reftosupp}}{\cref{supp:subsec:rgbd_version}}{Sec.~S.5.3} where we also compare against an RGB-D version of \ourName{}. 
}

{\parskip=1pt
\noindent
\textit{Detection Speed}:
Aside from a lower pose and bounding box error, \textit{\ourName{}} processes frames faster than \textit{A-SDF}. \cref{tab:detection_speed} shows a reduction in inference time of more than 60 times while still persevering the same level of detail.
}

%% file: content/assets/tables/canonical_reconstruction_small.tex
\begin{table}
    \centering
    \footnotesize
    \caption[]{Decoder Optimization Results. Each object is sampled in 50 different joint states for training as well as for testing. We average over all instances from our seven categories here. For a category-level comparison see \ifthenelse{\boolean{reftosupp}}{\cref{tab:decoder_results}}{Tab.~S.3}. \textit{$\dagger$ means the model is trained only on one category, thus the joint type prediction is not applicable (N/A). The joint state error mean is only reported across the revolute categories, as there is only one prismatic category (reported in brackets)}.}
    \label{tab:decoder_results_short}
    \begin{tabular}{l|ccc}
        \toprule
        Method & \makecell{CD ($\downarrow$)} & \makecell{Joint State\\Error ($\downarrow$)} & \makecell{Joint Type\\Accuracy ($\uparrow$)} \\
        \midrule
        A-SDF \cite{mu_a-sdf_2021} $\dagger$ & 1.437 & {11.337$\degree$} ($0.094\si{m}$) & N/A\\
        \ourName{} $\dagger$  & 1.190 & $12.474\degree$ (\textbf{$0.081\si{m}$}) & N/A\\
        A-SDF \cite{mu_a-sdf_2021} & \textbf{0.934} & $16.139\degree$ ($0.235\si{m}$) & \textbf{0.962}\\
        \ourName-No-Enf & 2.246 & $35.892\degree$ ($0.104\si{m}$) & 0.646 \\
        \ourName{} & 1.192 & \textbf{11.512$\degree$} ($0.141\si{m}$) & {0.908}\\
        \bottomrule
    \end{tabular}
    \vspace{-0.3cm}
\end{table}

%% file: content/6_conclusion.tex
\section{Conclusion}
\label{sec:conclusion}
We presented a novel method to reconstruct multiple articulated objects in a scene in a category- and joint-agnostic manner from a single stereo image. For reconstruction we learn a SDF-based decoder and show the necessity of regularization to achieve good performance. Our full single-shot pipeline improves over current two-stage approaches in terms of 3DIoU and inference speed.

{\parskip=5pt
\noindent\textbf{Limitations}:
While \ourName{} is able to generalize to unseen instances, it still relies on a learned shape prior. Using test time adaption techniques such as done by \cite{mu_a-sdf_2021} helps mitigating this issues, but is not sufficient to deal with categorically different objects. Additionally, while the single-forward pass is fast, jointly optimizing for pose, scale, codes like done in \cite{liu_catre_2022, irshad_2022_shapo} could further improve results with the cost of added execution time. Currently \ourName{} is only trained on objects with a single joint. To extend \ourName{} to objects with an arbitrary number of joints, we must be able to calculate pairwise similarity between two object states. While not explored in this paper, \ourName{} introduces a framework for future research to pursue this research question. A potential solution could leverage \cref{eqn:joint_sim_real} and Hungarian matching of the cross-product of articulation states to obtain similarities measurements between arbitrary kinematic structures.
}

%% file: content/supp/0_abstract.tex
In this supplementary material, we first describe our model architecture in more detail in \cref{supp:sec:model_architectures}. Next, we detail our backward optimization procedure in \cref{supp:sec:backward_optim}. In \cref{supp:sec:joint_code_space}, we visualize the difference between joint code spaces trained with and without our regularization and show the fitted polynomial functions, previously introduced in \ifthenelse{\boolean{reftomain}}{\cref{subsec:inverse_joint_decoder}}{Sec.~3.3.3}. Subsequently, we present example images from our synthetic and real dataset in \cref{supp:sec:datasets}. Lastly, in \cref{supp:sec:additional_metric}, we report extended results for our canonical reconstruction task in \ifthenelse{\boolean{reftomain}}{\cref{subsec:canonical_reconstruction_task}}{Sec.~4.2} and our full pipeline experiment in \ifthenelse{\boolean{reftomain}}{\cref{subsec:full_pipeline_task}}{Sec.~4.3}.

%% file: content/supp/A_architectures.tex
\section{Model Architecture}
\label{supp:sec:model_architectures}
\begin{figure*}
    \centering
    \includegraphics[width=\linewidth]{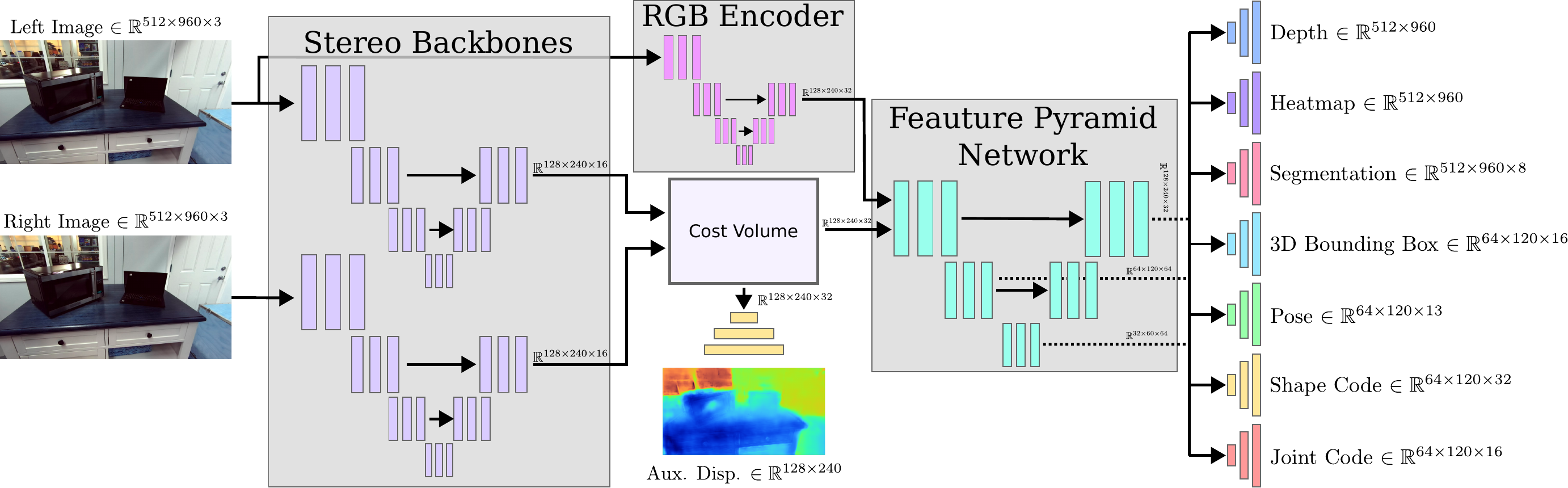}
    \caption{Encoder Architecture based on \cite{laskey2021simnet}}
    \label{supp:fig:encoder_architecture}
\end{figure*}

We present our model architecture for the encoder in \cref{supp:fig:encoder_architecture} and for the decoder in \cref{supp:fig:decoder_architecture}. 

\subsection{Encoder}
\label{supp:subsec:encoder_architecture}
{\parskip=0pt
Our encoder builds upon the SimNet-architecture \cite{laskey2021simnet}. The input is a stereo RGB image pair of size $\realspace^{960 \times 512 \times 3}$. Each image gets passed through a shared feature encoder network that outputs a low-dimensional feature map of size $\realspace^{128 \times 240 \times 16}$. This output is then fed into a cost volume which performs approximate stereo matching. Based on the result of the cost volume of size $\realspace^{128 \times 240 \times 32}$, a lightweight head predicts an auxiliary disparity map of size $\realspace^{128 \times 240}$.  
Parallel to that, we feed the left image through a separate RGB encoder that also predicts a feature map of size $\realspace^{128 \times 240 \times 32}$. This map, as well as the output of the cost volume, get concatenated and fed into a feature pyramid network which predicts three feature maps of sizes $\realspace^{128 \times 240 \times 32}$, $\realspace^{64 \times 120 \times 64}$, $\realspace^{32 \times 60 \times 64}$. Finally, using these features, each quantity described in \ifthenelse{\boolean{reftomain}}{\cref{subsec:encoder}}{Sec.~3.1} is predicted by its respective output head, including a segmentation mask, 3D bounding box, object pose, full resolution disparity, shape code, and joint code heads.

As in~\cite{laskey2021simnet}, although the stereo input for sim-to-real transfer has benefits for perceiving objects in harsh lighting conditions and for transparent or reflective objects, a RGB-D version 
could be trained as well (see \cref{supp:subsec:rgbd_version}).
}

\subsection{Decoder}
\label{supp:subsec:decoder_architecture}
Our decoder is split into two sub-decoders. A geometry decoder (see \ifthenelse{\boolean{reftomain}}{\cref{subsubsec:geometry_decoder}}{Sec~3.2.1}) based on DeepSDF \cite{park_deepsdf_2019} and a joint decoder (see \ifthenelse{\boolean{reftomain}}{\cref{subsubsec:joint_decoder}}{Sec.~3.2.2}). We detail both the architecture in the subsequent paragraphs.

\begin{figure}
    \centering
    \includegraphics[width=\linewidth]{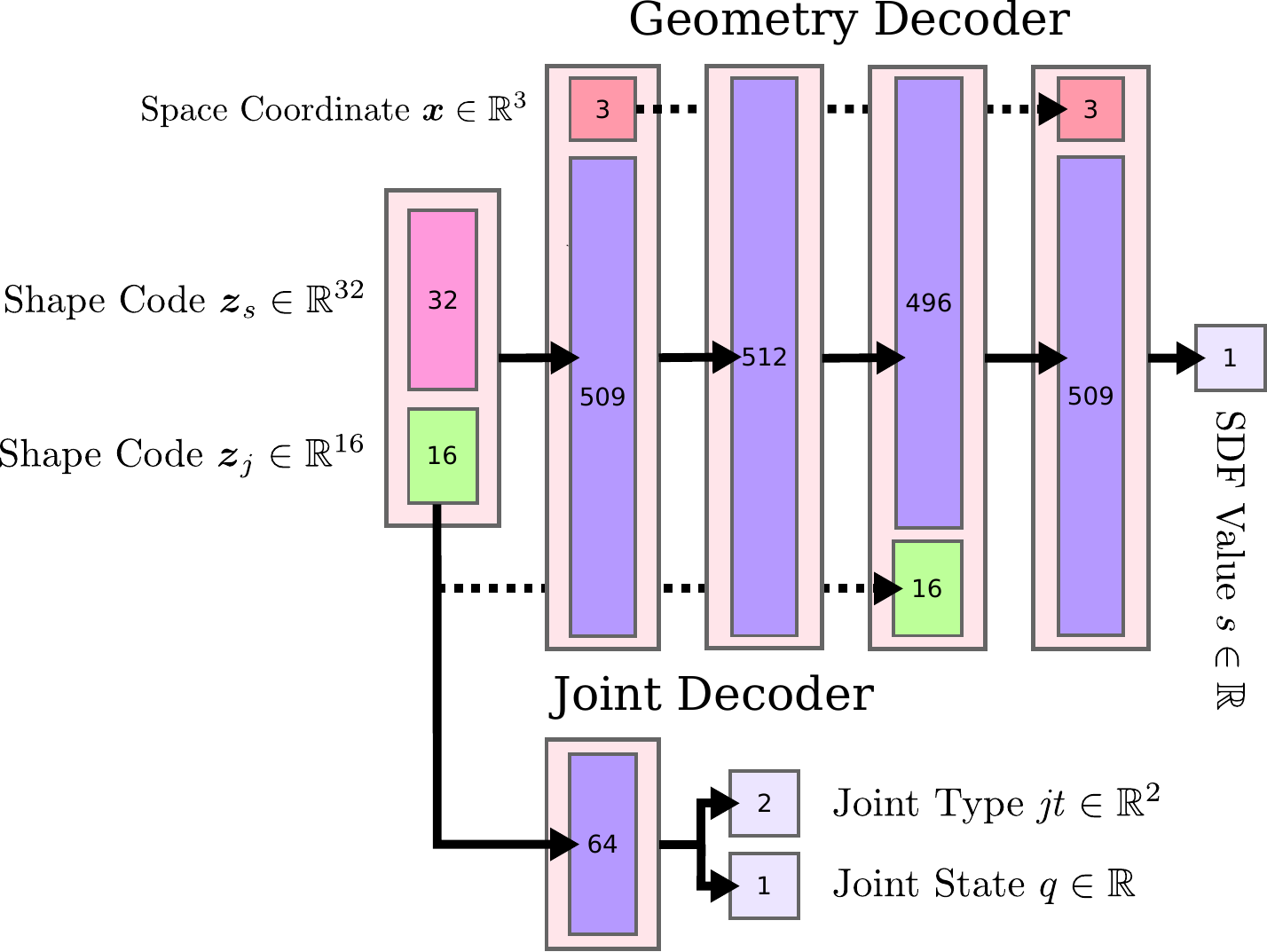}
    \caption{Decoder Architecture. The numbers indicate the size of the respective feature vector. Each arrow represents a layer of a multi-layer perceptron. For the geometry decoder, except for the first layer, the input to a layer always has a size of 512. The output dimensions vary depending on auxiliary inputs.}
    \label{supp:fig:decoder_architecture}
\end{figure}

\noindent \textbf{The geometry decoder} is a deep multi-layer perceptron consisting of four layers. The first layer takes a shape code $\latentShape$ and joint code $\latentJoint$ as input. Before the second and last layer, we concatenate the space coordinate $\spaceCoordinate$ for which we want to retrieve the SDF-value $\sdfValue$ with the output of the previous layer. Following the findings in \cite{mu_a-sdf_2021}, we again input the joint code $\latentJoint$ before the second last layer. For the exact feature vector dimensions see \cref{supp:fig:decoder_architecture}. As an activation function, we use ReLU for all except the last layer which uses tanh. 
Exploring exact input positions for shape code $\latentShape$, joint code $\latentJoint$, and space coordinate $\spaceCoordinate$, could be a topic of further research.

\noindent \textbf{The joint decoder} only takes a joint code $\latentJoint$ as input and feeds it through a single layer outputting a feature vector with 64 dimensions. This vector is then used to regress the articulation state, consisting of the continuous joint state $\jointState$ (no activation) and the discrete joint type $\jointType$ (Sigmoid activation).

An overview of the used loss scaling hyperparameters is given in \cref{supp:tab:decoder_scaling_training}.

\begin{table}
    \centering
    \begin{tabular}{lc}
        \toprule
        Scaling Variable & Value \\
        \midrule
        $\lossScaling_{\indexCodeRegularizer, \latentJoint, \text{pre}}$ & 0.1 \\
        $\lossScaling_{\indexJointRegularizer, \text{pre}}$ & 1.0 \\
        $\lossScaling_{\indexCodeRegularizer, \latentShape}$ & 0.0001 \\
        $\lossScaling_{\indexCodeRegularizer, \latentJoint}$ & 0.001 \\
        $\lossScaling_{\indexReconstruction}$ & 1.0 \\
        $\lossScaling_{\indexJointType}$ & 0.001 \\
        $\lossScaling_{\indexJointState}$  & 0.1 \\
        $\lossScaling_{\indexJointRegularizer}$ & 0.1 \\
        \bottomrule
    \end{tabular}
    \caption{Scaling Hyperparameters for Decoder Training.}
    \label{supp:tab:decoder_scaling_training}
\end{table}

%% file: content/supp/B_backward_optim.tex
\section{Backward Optimization}
\label{supp:sec:backward_optim}

The goal of the backward optimization is to retrieve the shape code $\latentShape^{u}$ and joint code $\latentJoint^{u}$ of an unknown object. The object is given through sampled SDF values, in total $P$. We denote the set of all $P$ SDF-space coordinate tuples as $\allSdfValues^{u} = \{(\spaceCoordinate_{p}, \sdfValue_{p})_{p} \forall p \in P \}$ for this unknown object. The problem can then be formalized as 
\begin{align}
    \label{supp:eqn:backward_optim}
    &\latentShape^{u}, \latentJoint^{u} = \\
    &\quad \arg\min_{\latentShape^{u}, \latentJoint^{u}} 
    \frac{1}{|\allSdfValues^{u}|} \sum_{(\spaceCoordinate_{p}, \sdfValue_{p}) \in \allSdfValues^{u}} \left| \geometryDecoder(\latentShape^{u}, \latentJoint^{u}, \spaceCoordinate_{p}) -  \sdfValue_{p} \right|, \nonumber
\end{align}
a minimization of the distance between the given SDF values and the one predicted by our geometry decoder (see \ifthenelse{\boolean{reftomain}}{\cref{subsubsec:geometry_decoder}}{Sec.~3.2.1}).

\begin{algorithm*}
    \caption{Backward Optimization: The goal is to retrieve shape and joint code for an unknown shape $\allSdfValues^{u}$ with $n$ hypotheses in parallel. Here, for clarity, we show the optimization for a single $i \in [1, \ldots, n]$. Eventually, from all returned code pairs the pair having the lowest distance error (see \cref{supp:eqn:backward_optim}) is returned. This procedure can be efficiently parallelized on a GPU.}
    \label{supp:algo:backward_optimization}
    \begin{algorithmic}[1] 
        \Procedure{BackOptimSingle}{$\allSdfValues^{u}, \boldsymbol{Z}_{\text{j}}, i$}
            \State $ \text{project}(\bullet), \text{reproject}(\bullet) \gets \text{SVD}(\boldsymbol{Z}_{\text{j}}) $
                \Comment{Retrieve project (see \cref{supp:joint_code_projection}) and reproject (see \cref{supp:joint_code_reprojection}) function}
            \State $\latentShape^{i} \gets \mathcal{N}( \boldsymbol{0}, \Sigma )$ 
                \Comment{Initialize shape codes with $\boldsymbol{0} \in \realspace^{\dimensionsShapeSpace}, \Sigma = \text{diag}(0.5) \in \realspace^{\dimensionsShapeSpace \times \dimensionsShapeSpace}$ }
            \State $\latentJoint^{i} \gets 
            \begin{cases}
                \bar{\boldsymbol{Z}}_{\text{j}}^{\text{prismatic}} & i \mod 2 = 0   \\ 
                \bar{\boldsymbol{Z}}_{\text{j}}^{\text{revolute}}  & i \mod 2 = 1
            \end{cases} $
            \Comment{Initialize joint codes}
            \State $\hat{\latentJoint}^{i} \gets \text{project} (\latentJoint^{i}) $
            \Comment{Project joint codes}
            \For{$\textit{step} \in [1, \ldots, 800$]}
                \State $\begin{aligned} 
                    \textit{loss}^{i} \gets &  \frac{1}{| \allSdfValues^{u} |} \sum_{(\spaceCoordinate_{p}, \sdfValue_{p}) \in \allSdfValues^{u}} \left| \geometryDecoder(\latentShape^{i}, \text{reproject}(\hat{\latentJoint}^{i}), \spaceCoordinate_{p}) -  \sdfValue_{p} \right|  
                    \\ 
                    & + 5 \cdot 10^{-3} ||\latentShape^{i}|| \\
                    & + 10^{-2} \min( ||\text{reproject}(\hat{\latentJoint}^{i}) - \boldsymbol{Z}_{\text{j}}||)  
                \end{aligned}$
                \Comment{Sum distance loss and regularization terms}
                \If{$\textit{step} \leq 600$}
                    \State $\latentShape^{i}, \hat{\latentJoint}^{i} \gets \text{ADAM}(\textit{loss}^{i})  $
                    \Comment{Update shape and joint code}
                \ElsIf{$600 < \textit{step} \leq 700$}
                    \State $\latentShape^{i} \gets \text{ADAM}(\textit{loss}^{i})  $
                    \Comment{Update shape code}
                \Else
                    \State $\hat{\latentJoint}^{i} \gets \text{ADAM}(\textit{loss}^{i})  $
                    \Comment{Update joint code}
                \EndIf
            \EndFor\label{optim}
            \State \textbf{return} $ \latentShape^{i}, \text{reproject}(\hat{\latentJoint}^{i}) $
        \EndProcedure
    \end{algorithmic}
\end{algorithm*}

At the beginning of the optimization, we randomly sample a set of 16 random shape codes from a zero-mean Gaussian distribution with a variance of $0.5$ as well as a set of 16 corresponding joint codes. For the joint codes, we do not sample but rather take the mean from all final joint codes of the training set after training $\latentJoint^{\jointIndex} \forall \jointIndex \in N$. We split the joint codes, where one half is using the mean of all prismatic training joint codes and the other half uses the mean of all revolute training joint codes. To guide the optimization through our latent joint code space, we propose a projection of the space as well as bounding the joint code variables.

{\noindent \textbf{SVD Projection}: To facilitate optimization along significant axes we will construct a projection based on the singular value decomposition of our training joint codes. To that end, we stack all training joint codes}
\begin{equation}
    \boldsymbol{Z}_{\text{j}} = \begin{bmatrix}
        \vdots\\
        {\latentJoint^{\jointIndex}}^\text{T}\\
        \vdots
    \end{bmatrix} \in \realspace^{\jointCount \times \dimensionsJointSpace}
\end{equation}
and do a singular value decomposition
\begin{align}
    &\boldsymbol{Z}_{\text{j}} - \bar{\boldsymbol{Z}_{\text{j}}} = \boldsymbol{U} \boldsymbol{\Sigma} \boldsymbol{V}^\text{T}, \\
    &\boldsymbol{U} \in \realspace^{\jointCount \times \jointCount}, \boldsymbol{\Sigma} \in \realspace^{\jointCount \times \dimensionsJointSpace}, \boldsymbol{V}^\text{T} \in \realspace^{\dimensionsJointSpace \times \dimensionsJointSpace}.
\end{align}
We then use 
\begin{equation}
    \label{supp:joint_code_projection}
    \hat{\latentJoint}^{u} = \left( \latentJoint^{u} - \bar{\boldsymbol{Z}_{\text{j}}} \right) \boldsymbol{V}
\end{equation}
to \textit{project} any joint code $\latentJoint^{u}$ and 
\begin{equation}
    \label{supp:joint_code_reprojection}
    \latentJoint^{u} = \hat{\latentJoint}^{u}\boldsymbol{V}^\text{T} + \bar{\boldsymbol{Z}_{\text{j}}}
\end{equation}
to \textit{reproject} a joint code $\hat{\latentJoint}^{u}$. 

We carry out the optimization from \cref{supp:eqn:backward_optim} in our projected space and thus, initially we project our joint codes using \cref{supp:joint_code_projection}. As well as in each optimization step, before inputting the joint code in our geometry decoder, we first reproject it using \cref{supp:joint_code_reprojection}. In initial testing, we found that this projection-reprojection step greatly helps navigate the high-dimensional space in which our joint codes reside in.

{\noindent \textbf{Bound Joint Code Variables}: On top of the previously described projection procedure, we ensure that the joint code variable is close to final joint codes from the training examples $\boldsymbol{Z}_{\text{j}}$ through minimizing the minimum distance to any joint code in the training set:}
\begin{equation}
    \min( ||\latentJoint^{u} - \boldsymbol{Z}_{\text{j}}||),
\end{equation}
where $||\bullet||$ is the row-wise Euclidean norm and $\min(\bullet)$ is a differentiable operator returning the minimum of a vector. An outline of our full optimization is presented in \cref{supp:algo:backward_optimization}.

%% file: content/supp/C_inverse_joint_decoder.tex
\section{Learned Joint Code Space}
\label{supp:sec:joint_code_space}
\begin{figure*}
    \centering
     \begin{subfigure}{\textwidth}
        \includegraphics[width=\textwidth]{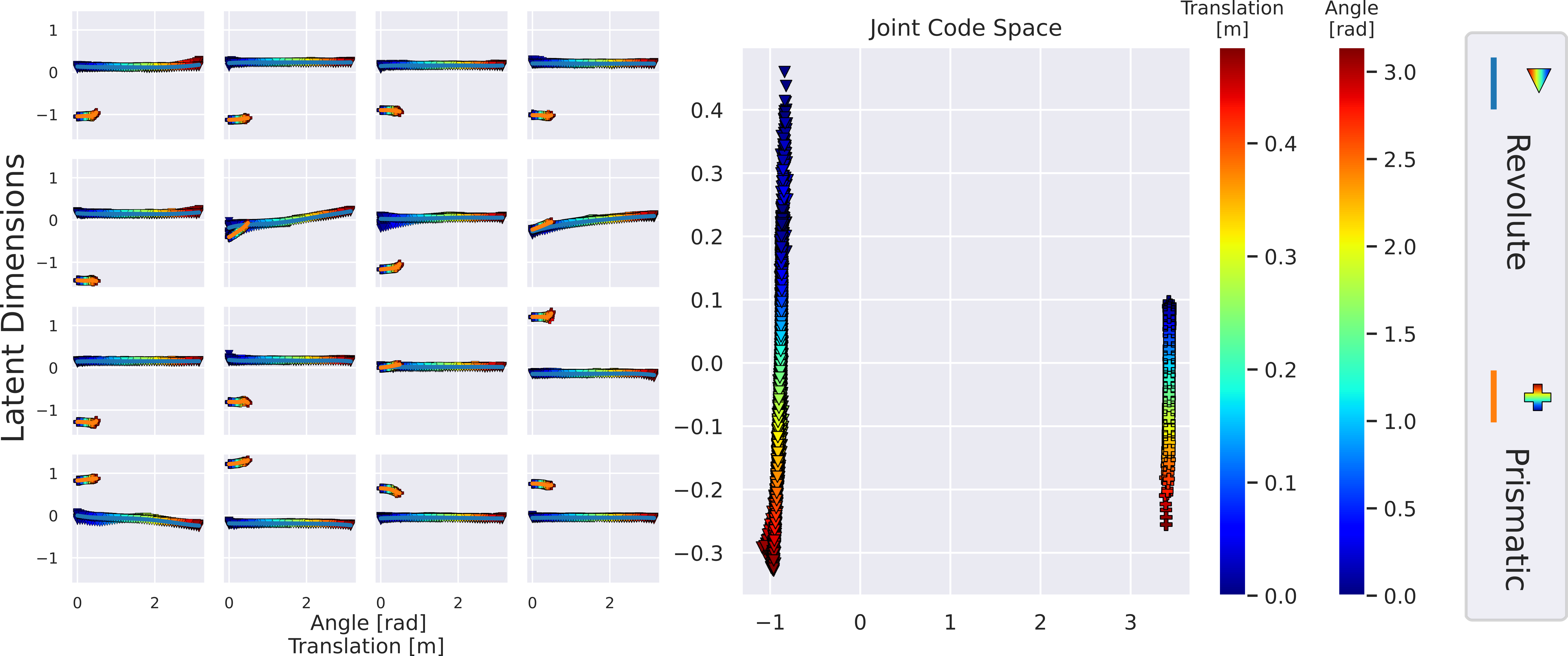}
        \caption{Using \ourName{}s Regularization.}
        \label{supp:subfig:carto_latent_joint_space}
    \end{subfigure}\\
    \begin{subfigure}{\textwidth}
        \includegraphics[width=\textwidth]{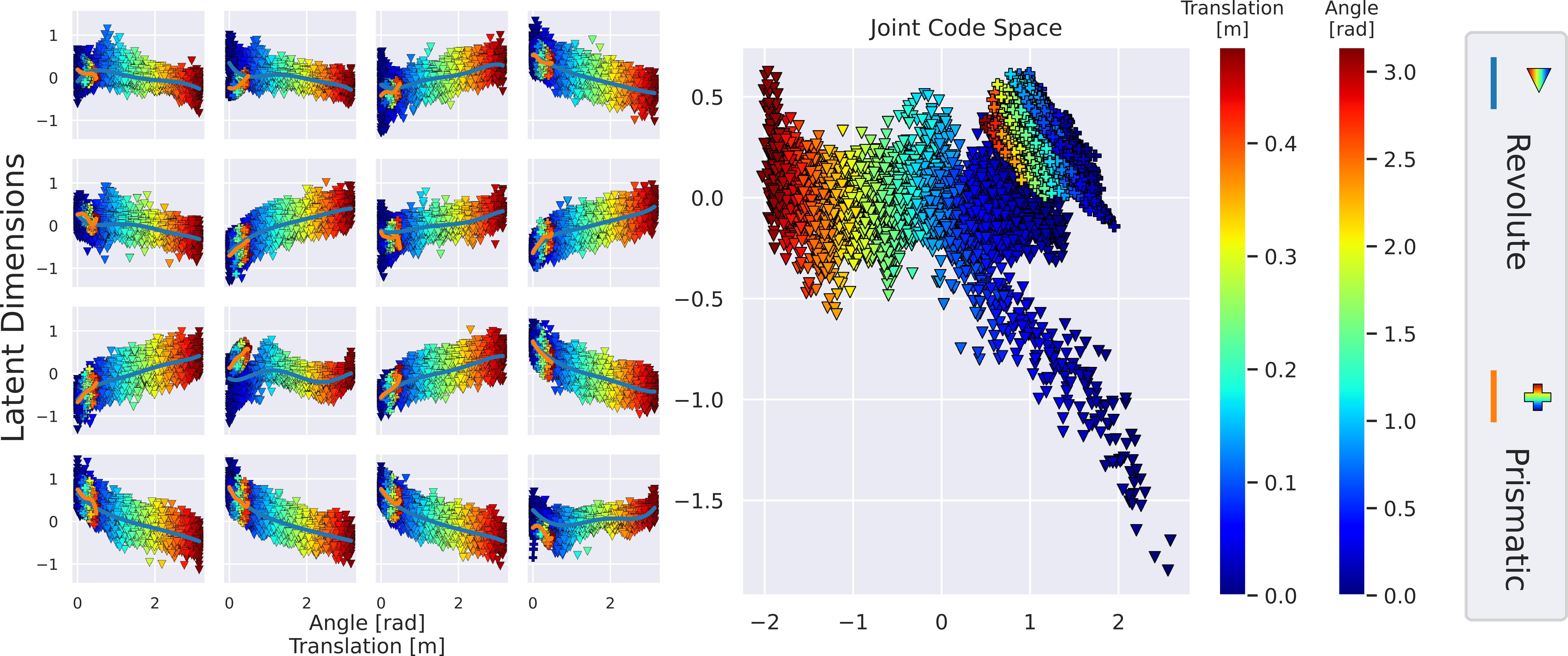}
        \caption{Only norm Regularization.}
    \end{subfigure}
    \caption[]{Comparison of Learned Joint Code Space. We compare the learned embedding of training joint codes when using our proposed regularization (a) against naively just regularizing the norm (b). An articulation state is expressed two-fold. First, by its form to represent the joint type. Here upside-down triangles stand for revolute and cross for prismatic joint types. Second, the form is colored by its joint state according to the scale shown on the right. In the left figure, each plot represents one component of the joint code $\latentJoint \in \realspace^{16}$. In the $i$-th plot, we plot the $i$-th component of all training joint codes on the y-axis against their associated known joint state on the x-axis. Additionally, we overlay the in 
    \ifthenelse{\boolean{reftomain}}{\cref{subsec:inverse_joint_decoder}}{Sec.~3.3.3} explained polynomial functions. In the right figure, we show a two-dimensional projection based on singular value decomposition of all training joint codes.}
    \label{supp:fig:latent_joint_spaces}
\end{figure*}

{\parskip=0pt
In this section, we visualize and compare the resulting learned latent joint space using our in \ifthenelse{\boolean{reftomain}}{\cref{subsubsec:joint_space}}{Sec.~3.3.1} introduced regularization against naively training it. In \cref{supp:fig:latent_joint_spaces}, we visualize the learned joint codes of the training results for \textit{\ourName} and \textit{\ourName-No-Enf} from \ifthenelse{\boolean{reftomain}}{\cref{subsec:canonical_reconstruction_task}}{Sec.~4.2}. When comparing both visualizations, we can explain the worse performance of \textit{\ourName-No-Enf} in \cref{tab:decoder_results}. 

\ourName{} trained without regularization struggles to correctly align the spaces such that joint codes belonging to the same articulation state, independent of the object, are close and show a low variance. The decoder rather learns to represent the final geometry of the articulated object jointly through both codes, the shape code $\latentShape$ and joint code $\latentShape$ instead of disentangling one from the other. One could argue that this case is similar to not splitting the codes. Compared to that, when using our proposed regularization, we learn a cleaner disentanglement between the shape and the articulation state of the object. Joint codes of similar articulation states in the training set are arranged closer in the latent joint embedding and thus, the variance in the y-direction across all plots on the left side of \cref{supp:fig:latent_joint_spaces} (a) is much lower when compared to training without our regularization in \cref{supp:fig:latent_joint_spaces} (b). Moreover, two distinct clusters are visible (prismatic and revolute) whereas without \ourName{}s regularization different joint types overlap. 
}

%% file: content/supp/D_datasets.tex
\section{Datasets}
\label{supp:sec:datasets}
\begin{figure*}[ht!]
    \centering
    \includegraphics[width=0.95\linewidth]{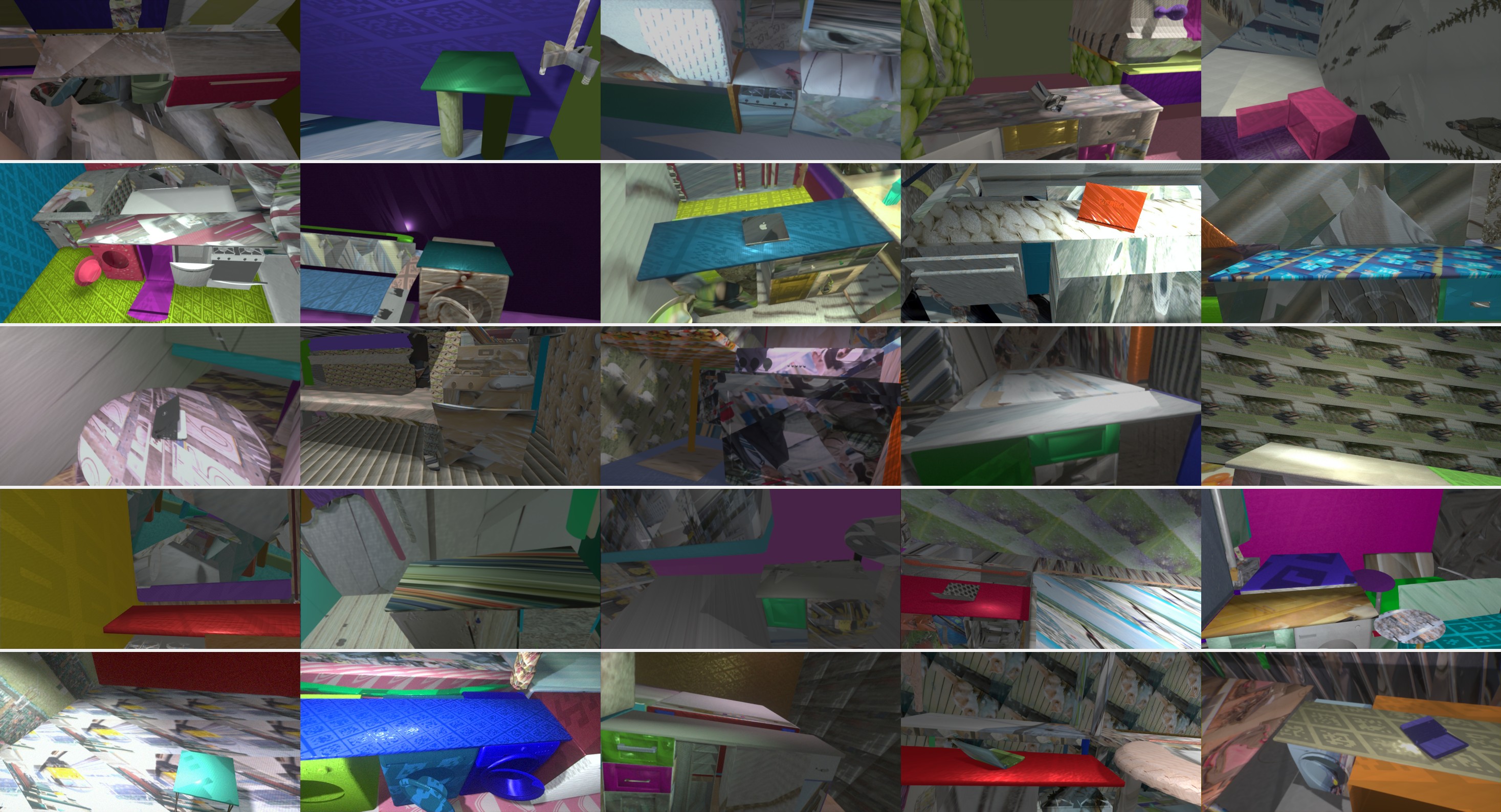}
    \caption{Synthetic Example Images}
    \label{supp:fig:syn_images}
    \vspace{1cm}
\end{figure*}

\begin{figure*}[htb!]
    \centering
    \includegraphics[width=0.95\linewidth]{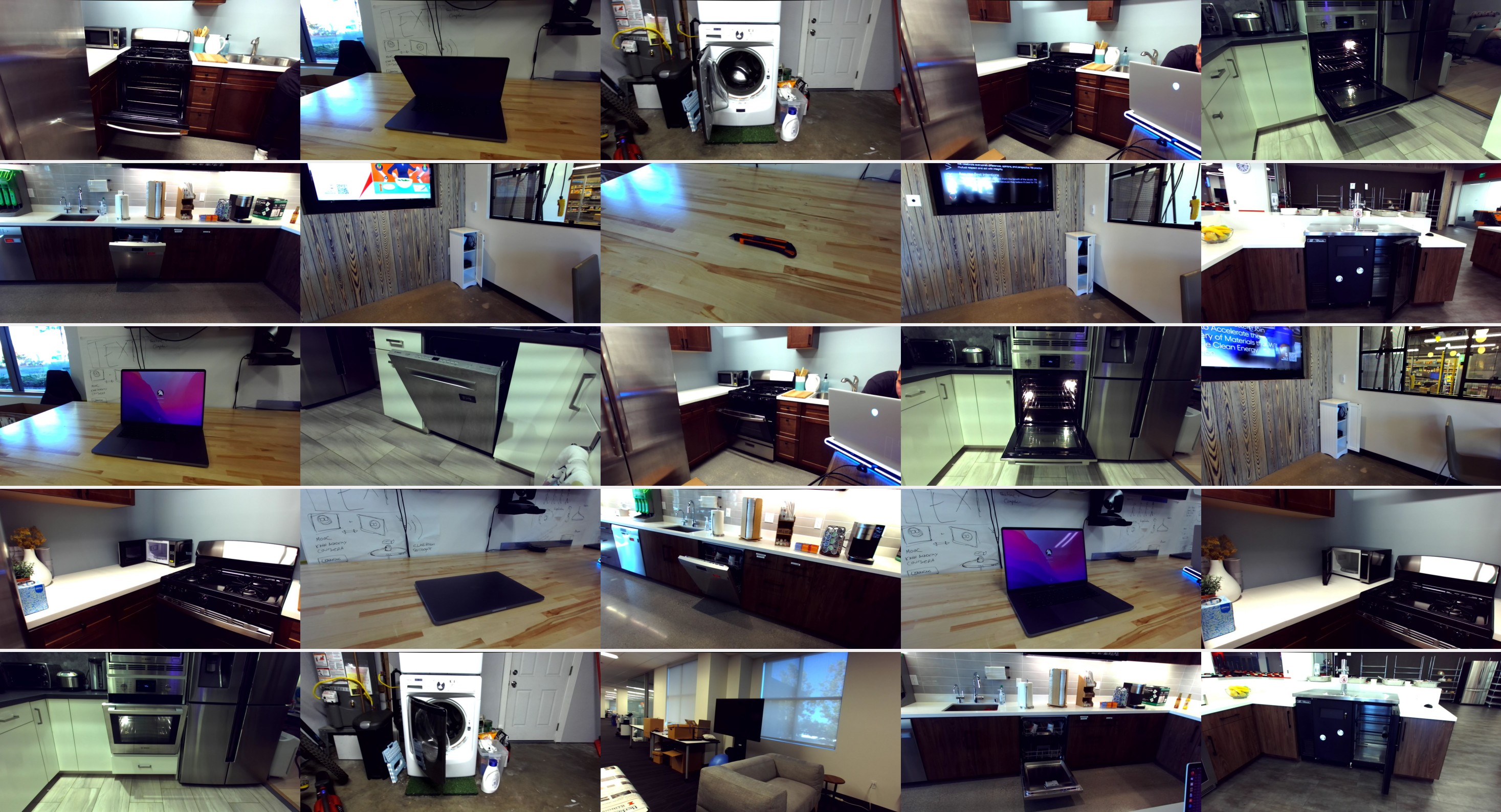}
    \caption{Real Example Images}
    \label{supp:fig:real_images}
\end{figure*}

{\parskip=0pt
\cref{supp:fig:syn_images} presents exemplary images of our procedural generated kitchen dataset. In total, we collected roughly 100k images for training and 20k images for testing. Due to the long run-time of our A-SDF baseline, we only evaluated on the first 2000 samples in which according to our ground truth objects are present. See \cref{supp:fig:real_images} for exemplary images from our collected real-world dataset.

In \cref{tab:dataset_comparison} we compare our dataset against the RBO \cite{martin-martin_rbo_2018} and the BMVC \cite{michel_pose_2015} dataset.
}

\input{content/assets/tables/dataset_comparison}

%% file: content/assets/tables/dataset_comparison.tex
\begin{table*}
    \centering
    \footnotesize
    \begin{tabular}{c|cccccc}
        \toprule 
        Dataset & \makecell{Categories available\\as 3D Models \textit{(All)}} & \makecell{Instances\\per Category} & \makecell{Total Unique Scenes} & \makecell{Input\\Modality} & Realism & \makecell{Corresponding\\Synthetic Images}\\
        \midrule
        RBO \cite{martin-martin_rbo_2018} & 4 \textit{(14)} & 1 & \textbf{385} & {RGB-D} & Lab + Human Occlusion & No\\ 
        BMVC \cite{michel_pose_2015} & 2 \textit{(3)} & 1 & 8 & {RGB-D} & \textbf{Real Environments} & No\\
        Ours & \textbf{7} \textit{(7)} & \textbf{2}* & 9 & \textbf{Stereo-RGB + {RGB-D}}  & \textbf{Real Environments} & \textbf{Yes}\\
        \bottomrule
    \end{tabular}
    \caption{\footnotesize Full Pipeline Dataset Comparison. While RBO \cite{martin-martin_rbo_2018} and BMVC \cite{michel_pose_2015} contain additional categories, these categories lack corresponding 3D meshes in public datasets such as PartNetMobility \cite{Xiang_2020_SAPIEN}. \textit{* one washing machine instance}}
    \label{tab:dataset_comparison}
\end{table*}

%% file: content/supp/E_additional_results.tex
\section{Additional Experimental Results}
\label{supp:sec:additional_metric}

\input{content/assets/tables/canonical_reconstruction_full}

\begin{table*}
    \small
    \caption{Reconstruction and Articulation State Prediction Results when using A-SDF \cite{mu_a-sdf_2021} with the Proposed Test-Time-Adaptation. \textit{$^*$The joint state error mean is only reported across the revolute categories, as there is only one prismatic category.}}    
    \label{supp:tab:asdf_tta}
    \centering
    \footnotesize
    \begin{tabular}{l|ccccccc|c|c}
        \toprule
        Method & Dishwasher & Laptop & Microwave & Oven & \makecell{Refrigerator} & Table & \makecell{Washing\\Machine} & \makecell{Instance\\Mean$^*$} & \makecell{Category\\Mean$^*$} \\
        \midrule
        Chamfer Distance ($\downarrow$) & 0.101 & 1.035 & 0.529 & 0.451 & 1.383 & 64.097 & 0.332 & 13.339 & 9.704 \\
        Joint State Error ($\downarrow$) & $19.387\degree$ & $18.675\degree$ & $58.088\degree$ & $25.432\degree$ & $20.700\degree$ & $0.552\si{m}$ & $51.467\degree$ & $29.024\degree$ & $32.292\degree$ \\
        \bottomrule
    \end{tabular}
\end{table*}

In this section, we present additional metrics for our experiments. Namely, using A-SDFs proposed test time adaption as well as the Chamfer distance and joint state error for the full pipeline experiment. Also, in addition to \ifthenelse{\boolean{reftomain}}{\cref{tab:decoder_results_short}}{Tab.~2}, we report the more fine-grained category-level metrics in \cref{tab:decoder_results}.

\subsection{Canonical Reconstruction Task: A-SDF TTA}
\label{supp:subsec:a-sdf_tta}
{\parskip=0pt
In our experiments (see \ifthenelse{\boolean{reftomain}}{\cref{subsec:canonical_reconstruction_task}}{Sec.~4.2}), the proposed test time adaptation (TTA) \cite{mu_a-sdf_2021} did not prove to be stable. We report the results in \cref{supp:tab:asdf_tta}. While for some object instances TTA reduces the Chamfer distance, for the table category TTA does not prove to be robust. Additionally, the joint state error increases substantially. Both behaviors are reasonable when reflecting on the proposed TTA. When jointly optimizing the input shape code, the joint state, and network weights, the entire network will overfit to the single given geometry. Thus, it is easier to achieve a lower Chamfer distance. Whereas, the joint state variable becomes unbound from other examples and can be optimized freely, losing its meaning and therefore, potentially resulting in a high joint state error.

The proposed TTA is still promising and with further investigation into how to mitigate the aforementioned problems, it can prove to be an ideal tool for reconstructing (articulated) objects in the wild \cite{irshad_2022_shapo}.
}

\subsection{Extended Metrics for Full Pipeline}
\label{supp:subsec:extend_metrics}
\begin{figure*}[b]
    \centering
    \begin{subfigure}{0.85\textwidth}
        \includegraphics[width=\textwidth]{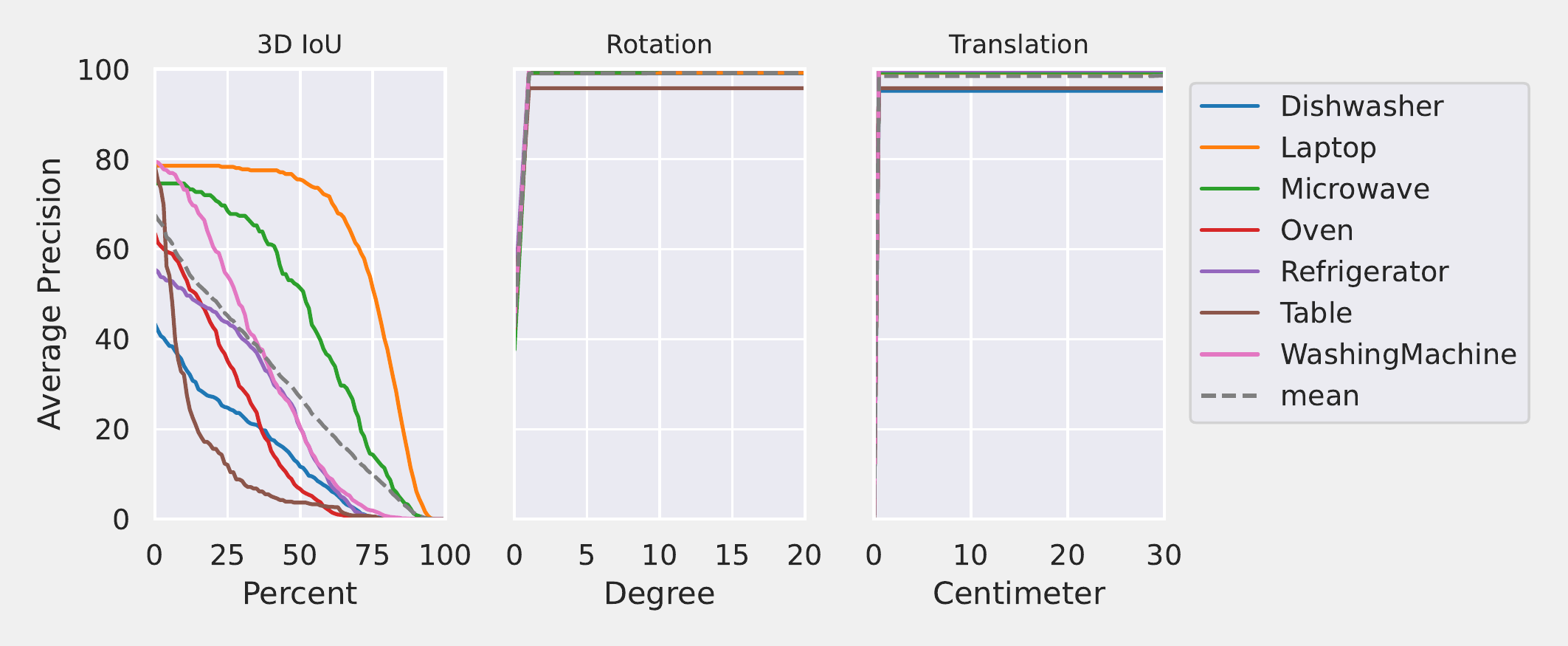}
        \caption{A-SDF-GT}
    \end{subfigure}\\
    \begin{subfigure}{0.85\textwidth}
        \includegraphics[width=\textwidth]{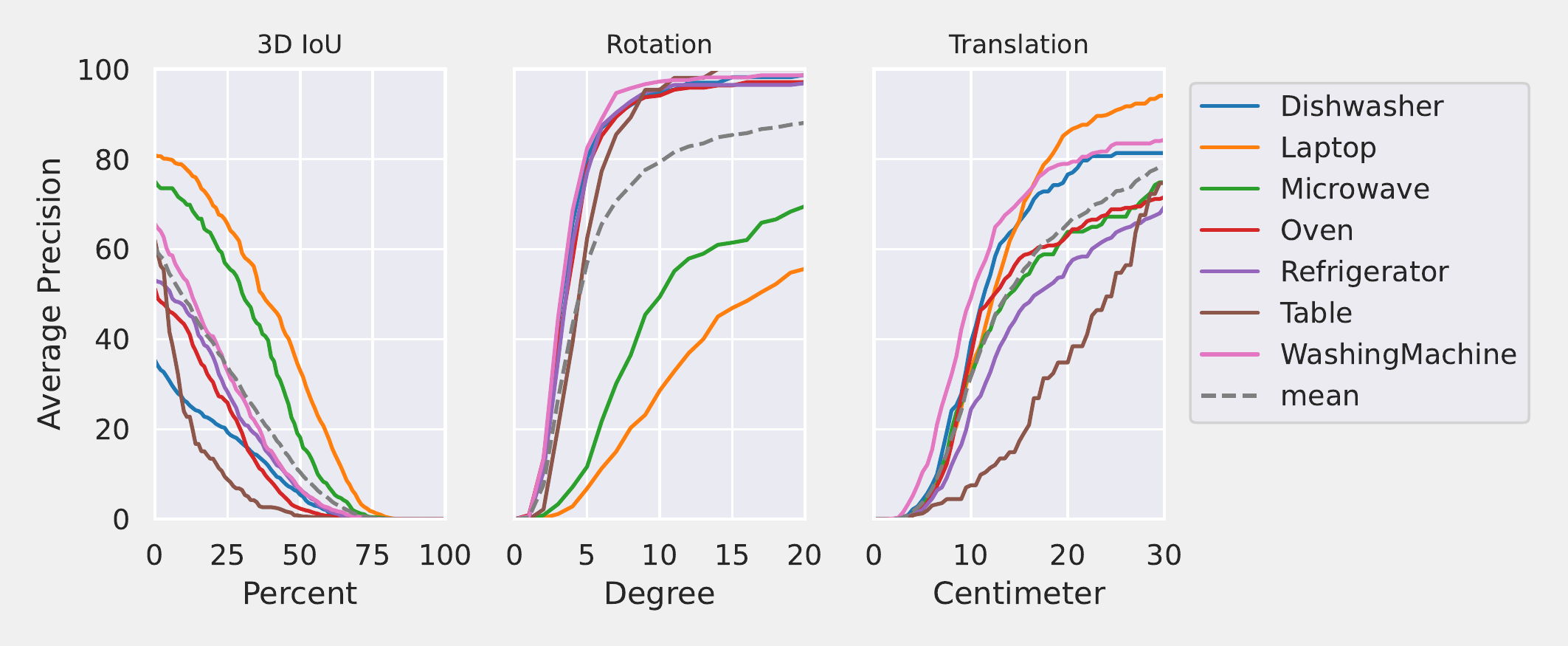}
        \caption{A-SDF}
    \end{subfigure}\\
    \begin{subfigure}{0.85\textwidth}
        \includegraphics[width=\textwidth]{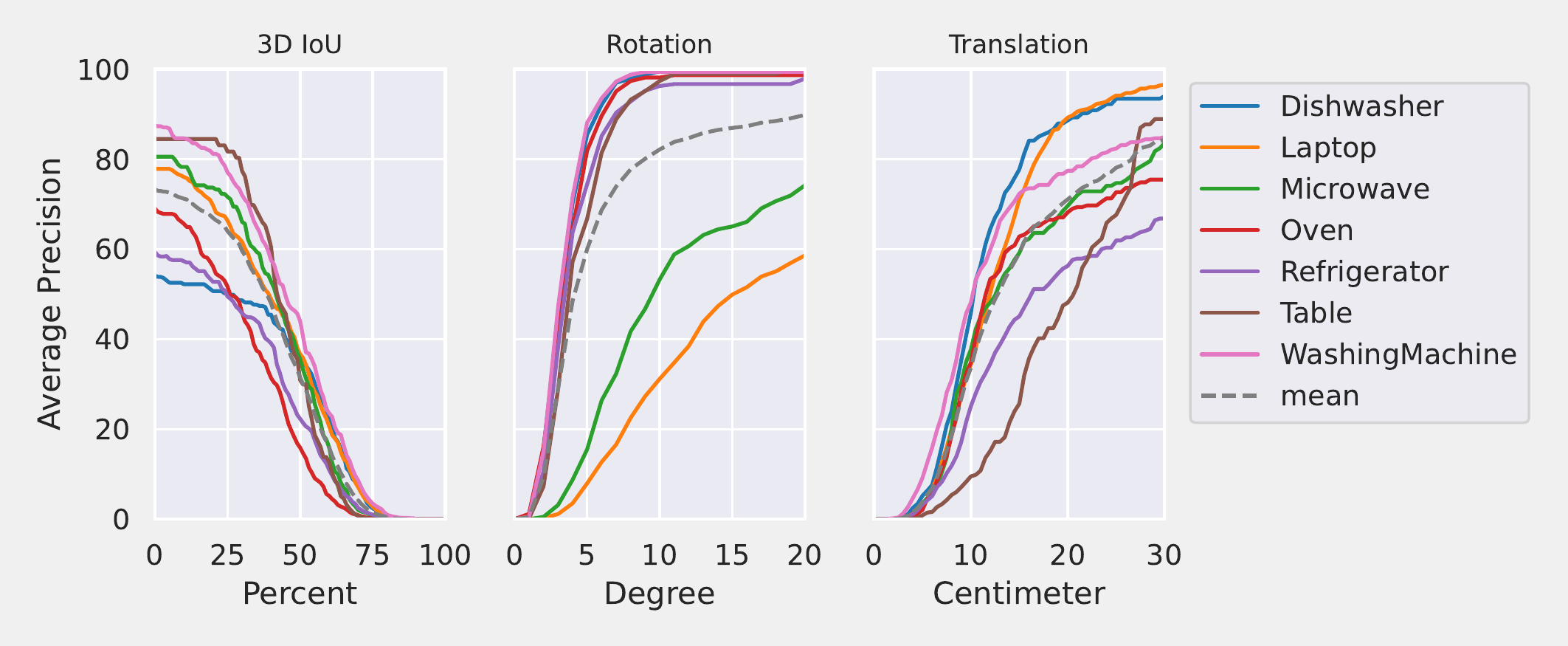}
        \caption{\ourName}
    \end{subfigure}
    \caption[]{Detailed metrics for the experiment presented in \ifthenelse{\boolean{reftomain}}{\cref{subsec:full_pipeline_task}}{Sec.~4.3}. We report the average precision for the 3D IoU and the pose prediction for each category in our test set as well as the mean over all instances. It can be observed that overall the mean 3D IoU is lower for \textit{\ourName{}} compared to \textit{A-SDF-GT} and \textit{A-SDF}. For \textit{A-SDF-GT} the laptop and microwave category stands out as they are mostly placed on counters and thus they are less occluded than other objects. As expected, the poses predicted by \textit{A-SDF} and \textit{\ourName{}} are similar as they both use the same pose map predicted by our encoder.}
    \label{supp:fig:3d_metrics}
\end{figure*}

\begin{figure*}[b]
    \centering
    \begin{subfigure}{0.85\textwidth}
        \includegraphics[width=\textwidth]{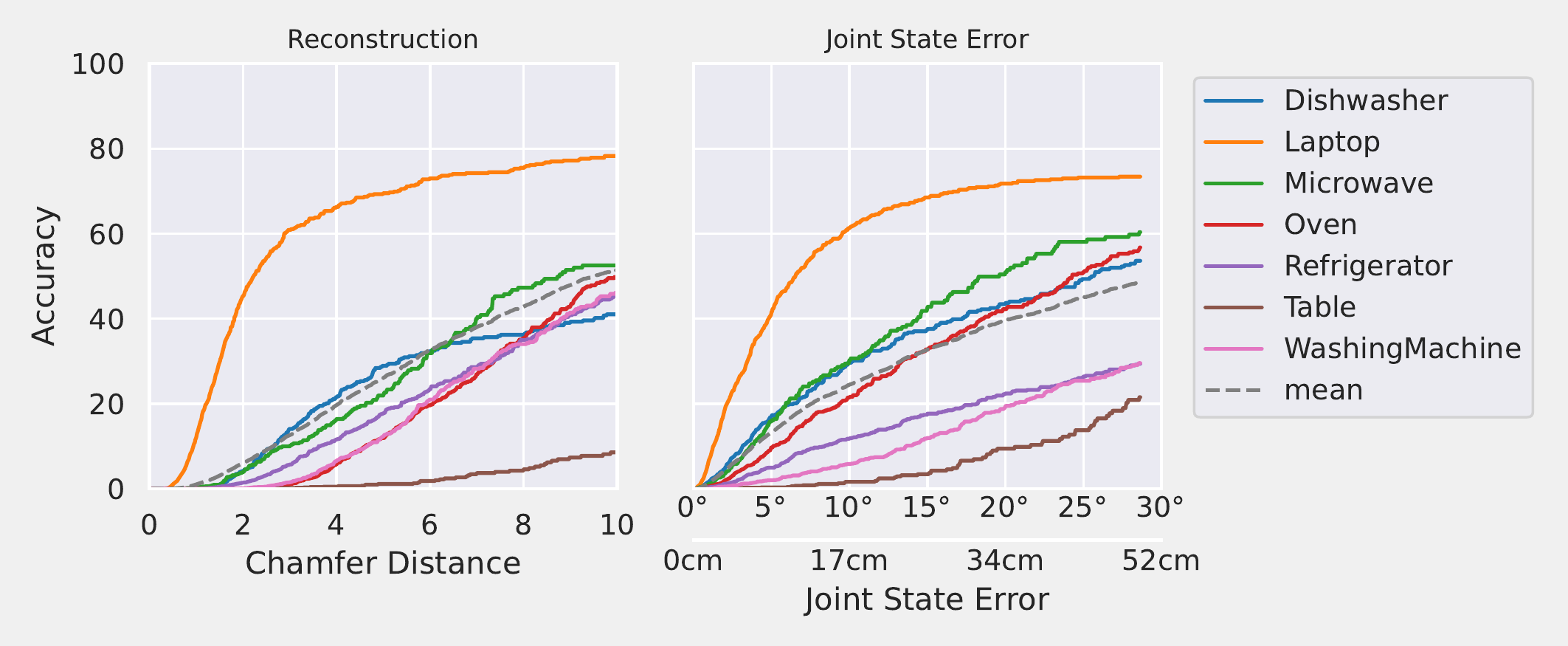}
        \caption{A-SDF-GT}
    \end{subfigure}\\
    \begin{subfigure}{0.85\textwidth}
        \includegraphics[width=\textwidth]{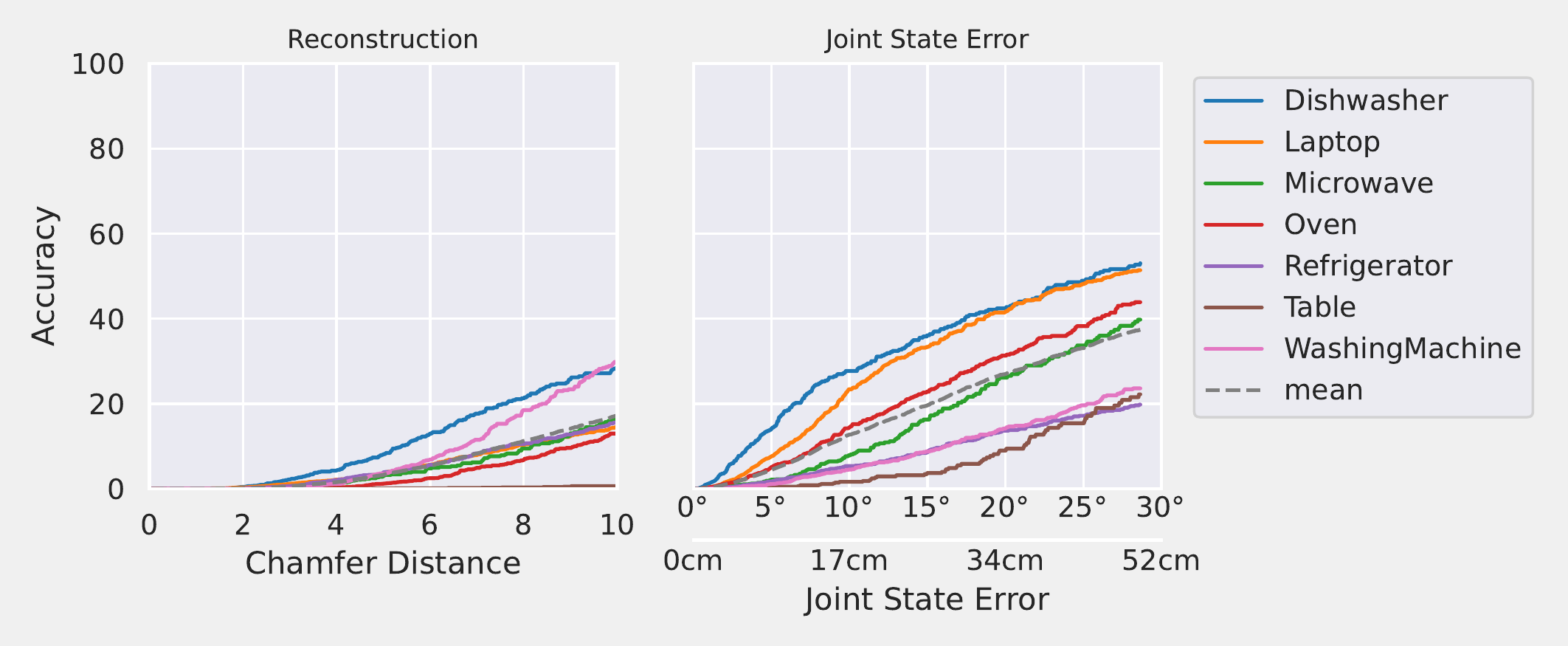}
        \caption{A-SDF}
    \end{subfigure}\\
    \begin{subfigure}{0.85\textwidth}
        \includegraphics[width=\textwidth]{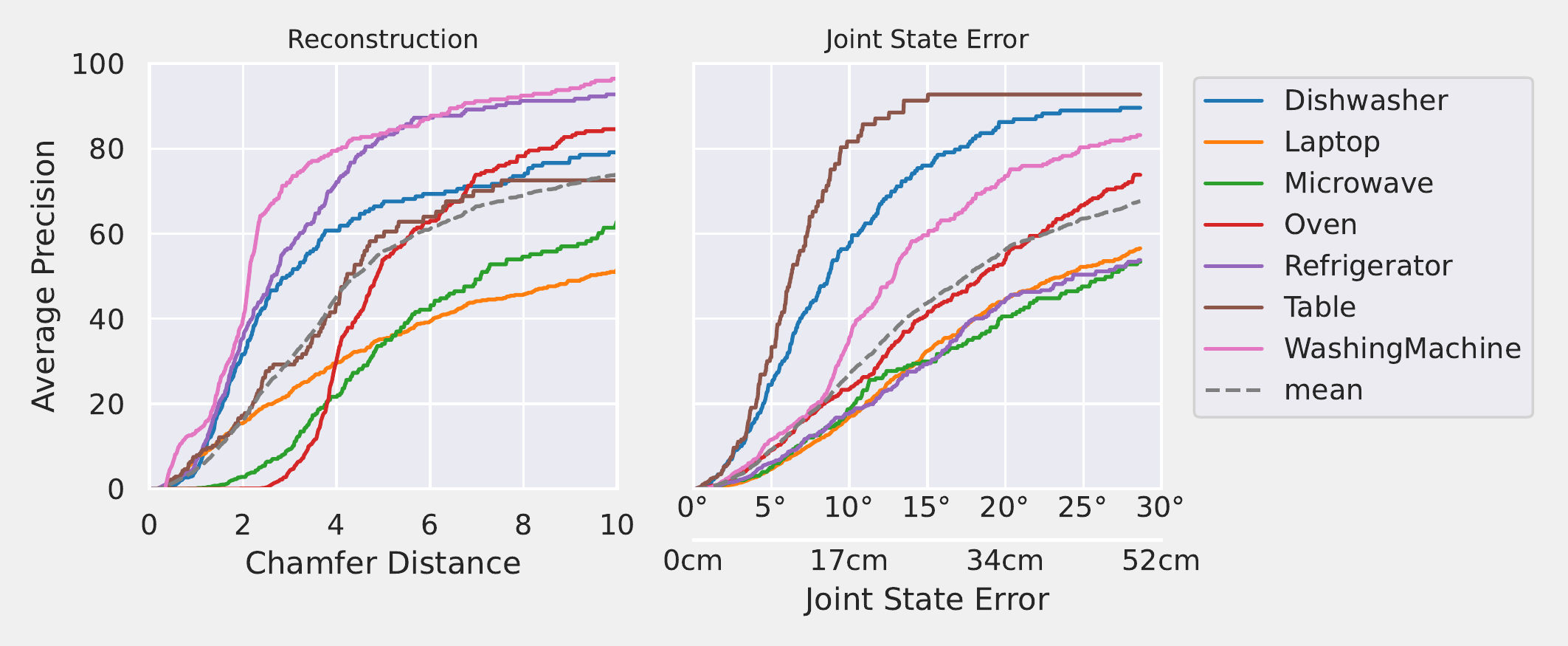}
        \caption{\ourName{}}
    \end{subfigure}
    \caption[]{Additional metrics for the experiment presented in \ifthenelse{\boolean{reftomain}}{\cref{subsec:full_pipeline_task}}{Sec.~4.3}. In addition to the metrics already reported in \ifthenelse{\boolean{reftomain}}{\cref{subsec:full_pipeline_task}}{Sec.~4.3}, we report the more fine-grained object-centric Chamfer distance as well as the joint state prediction error. Both metrics show a similar trend as the more coarse 3D IoU. One can observe though, that for the laptop category \textit{A-SDF-GT} performs significantly better than all other categories. Compared to that \textit{A-SDF}, which uses a predicted segmentation masks, does not show this special behavior for laptops. As laptops are small and the segmentation mask is very thin, this gap in performance highlights potential failure cases of an optimization-based method due to imperfect segmentation masks.}
    \label{supp:fig:chamfer_joint_metrics}
\end{figure*}

{\parskip=0pt
In addition to the tabular values reported in \ifthenelse{\boolean{reftomain}}{\cref{tab:full_pipeline_synthetic}}{Tab.~3a}, we present the respective mAP curve in \cref{supp:fig:3d_metrics}. The results highlight even more that an optimization-based two-stage approach suffers from its partial input. The two counter objects, laptops and microwaves, which are free-standing and thus much more points for reconstruction are available get reconstructed much better compared to other objects. On the other hand, for these objects, the predicted rotation is much worse. This can be rooted in the fact that for all other objects, we can learn a strong prior of the rotation being roughly camera facing, whereas, for laptops and microwaves the range of the possible rotation is much higher as they are placed freely on top of the counter.

While in \ifthenelse{\boolean{reftomain}}{\cref{tab:full_pipeline_synthetic}}{Tab.~3a} and previously we only discussed the overall 3D IoU and pose error, which gives a holistic evaluation of the full pipeline, we additionally report object-centric L2-Chamfer distances (similar to \cite{irshad_centersnap_2022, irshad_2022_shapo}) multiplied by $10^3$, as well as the joint state error in \cref{supp:fig:chamfer_joint_metrics}. Since this is an object-centric evaluation and should not evaluate the detection quality, we are very forgiving in selecting our detection matches. For each scene, we calculate our spatial 2D detections and retrieve the ground-truth spatial 2D detections from the heatmap, we then match the predicted and ground truth detections by solving a linear sum assignment problem, ignoring unmatched detections (either ground-truth or predicted). For each matched detection we then reconstruct the object as before using our geometry decoder and retrieve the joint through our joint decoder. We then calculate the Chamfer distance between the predicted points and the ground-truth points and compare the joint states.

In this experiment, we observe the same trend as for 3D IoU. One major difference is that \textit{A-SDF-GT} reconstructs laptops more accurately compared to \textit{A-SDF} and \textit{\ourName} which can be attributed to laptops having the least occlusion (either through self-occlusion or other objects).
}

\subsection{\ourName{} RGB-D Version}
\label{supp:subsec:rgbd_version}
\begin{figure*}[b]
    \centering
    \includegraphics[width=\textwidth]{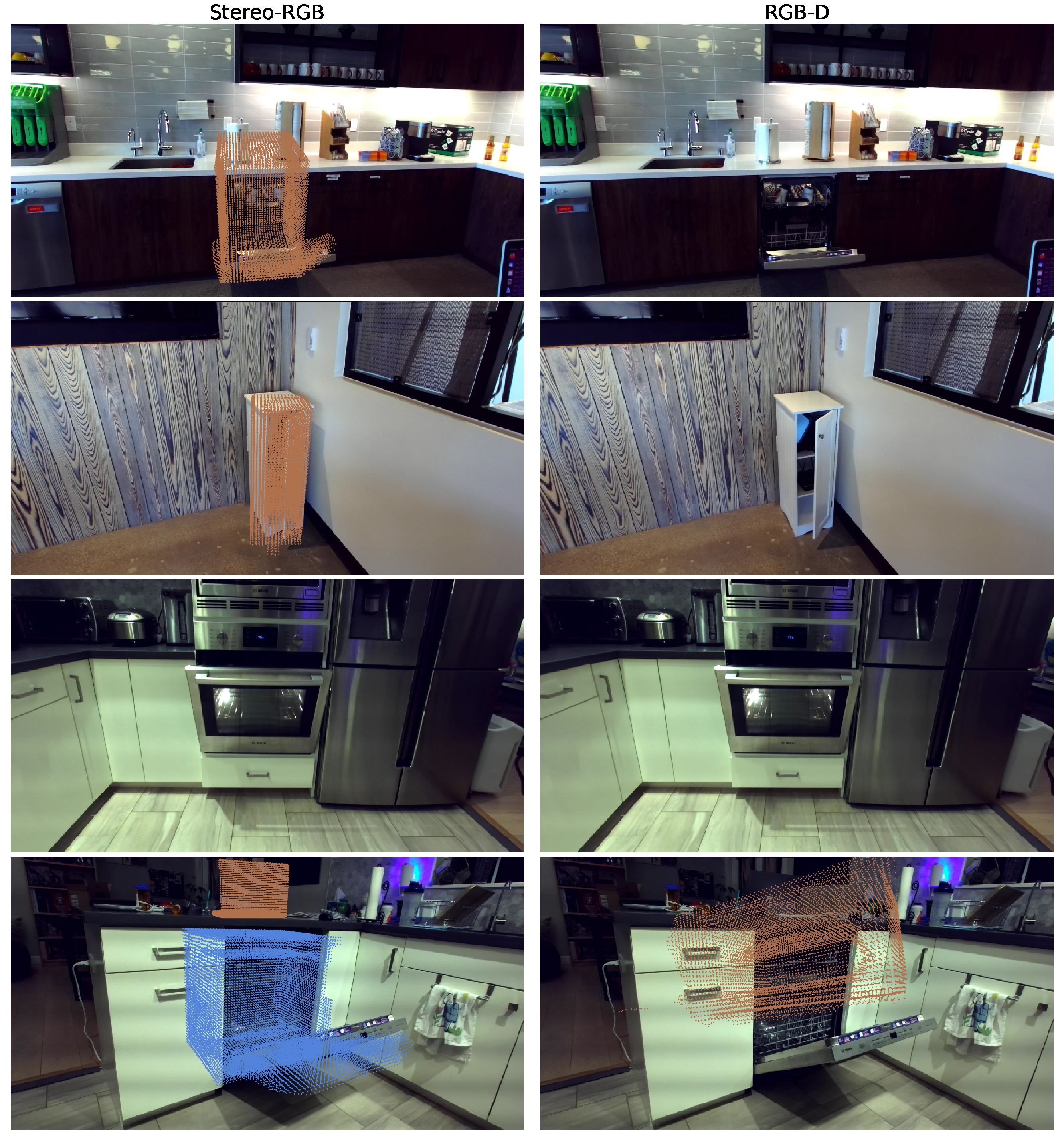}
    \caption{Stereo-RGB Image (Left) vs. RGB-D Image (Right) Input. The first row shows a successful detection and reconstruction of \ourName{} in an office kitchen environment. Second row shows a reconstruction of a cabinet. Eventhough, \ourName{} has never seen objects from this category it highlight its generalization beyond the trained categories. The third and fourth row show two failure cases of either no detection at all (third row) or a misdetection of a laptop on the kitchen counter (fourth row). CARTO with RGB-D input is not able to reconstruct any objects.}
    \label{supp:fig:stereo_rgbd_comparison}
\end{figure*}

{\parskip=0pt
In addition to the proposed stereo-RGB input version of \textit{\ourName{}}, we also evaluated and tested an RGB-D version \textit{\ourName{}-D}. 
We report quantitative results on the same synthetic dataset in \cref{supp:tab:stereo_rgbd} as well as compare the detections qualitatively in \cref{supp:fig:stereo_rgbd_comparison}. 

Quantitatively, the \textit{\ourName{}-D} performs slightly better compared to our proposed stereo RGB version. This is to be expected given that \ourName{} needs to learn the notion of depth first whereas \textit{\ourName{}-D} does not. Contrary to this observation, in our real world experiments, we do not get a single meaningful detection using the RGB-D input version (see \cref{supp:fig:stereo_rgbd_comparison}). Thus, overall, we decided for the proposed stereo version of \ourName{}.
}
\begin{table}
    \centering
    \footnotesize
    \caption{Full Scene Reconstructions Results with RGB-D Input.}
    \label{supp:tab:stereo_rgbd}
    \begin{tabular}{l|cccc}
        \toprule
            Method & IOU25 $\uparrow$ & IOU50 $\uparrow$ & $10\si{\degree}10\si{cm} \uparrow$ & $20\si{\degree}30\si{cm} \uparrow$  \\
        \midrule
            \ourName & {64.0} & {31.5} & \textbf{28.7} & {76.6} \\
            \ourName-D & \textbf{67.8} & \textbf{38.2} & {27.0} & \textbf{84.7} \\
        \bottomrule
    \end{tabular}
\end{table}%

%% file: content/assets/tables/canonical_reconstruction_full.tex
\begin{table*}
    \centering
    \caption{Decoder Optimization Results. Each object is sampled in 50 different joint states for training as well as for testing. \textit{$\dagger$ means the model is trained only on a single category.}}
    \label{tab:decoder_results}
    \begin{subtable}[h]{1.0\textwidth}
        \small
        \centering
        \footnotesize
        \begin{tabular}{l|ccccccc|c|c}
            \toprule
            Method & Dishwasher & Laptop & Microwave & Oven & \makecell{Refrigerator} & Table & \makecell{Washing\\Machine} & \makecell{Instance\\Mean} & \makecell{Category\\Mean} \\
            \midrule
            A-SDF \cite{mu_a-sdf_2021} $\dagger$  & 2.162 & {0.264} & 2.256 & 1.540 & 1.409 & 1.465 & 0.960 & 1.437 & 1.418 \\
            \ourName{} $\dagger$  & 1.874 & 0.965 & 1.429 & 1.606 & {0.690} & {1.330} & {0.437} & 1.190 & 1.252\\
            A-SDF \cite{mu_a-sdf_2021} & \textbf{0.336} & \textbf{0.601} & \textbf{0.700} & \textbf{0.883} & 1.425 & 1.862 & \textbf{0.608} & \textbf{0.934} & \textbf{0.916} \\
            \ourName-No-Enf & 1.043 & 3.820 & 2.685 & 2.317 & 2.454 & \textbf{1.676} & 1.727 & 2.246 & 2.248\\
            \ourName{} & 0.554 & 1.448 & 0.782 & 2.056 & \textbf{0.988} & 1.688 & 0.830 & 1.192 & 1.181\\
            \bottomrule
        \end{tabular}
        \caption{Shape Reconstructions Results. We report the bi-directional L2-Chamfer distance (CD) ($\downarrow$) times 1000 between the original mesh and the reconstructed version.}
        \label{tab:shape_reconstruction_results}
    \end{subtable}%
    \newline
    \vspace*{0.2cm}
    \newline
    \begin{subtable}[h]{0.95\textwidth}
        \small
        \centering
        \footnotesize
        \begin{tabular}{l|ccccccc|c|c}
            \toprule
            Method & Dishwasher & Laptop & Microwave & Oven & \makecell{Refrigerator} & Table & \makecell{Washing\\Machine} & \makecell{Instance\\Mean$^*$} & \makecell{Category\\Mean$^*$} \\
            \midrule
            A-SDF \cite{mu_a-sdf_2021} $\dagger$  & {1.616$\degree$} & $17.282\degree$ & $11.161\degree$ & {4.045$\degree$} & {19.254$\degree$} & $0.094\si{m}$ & $16.462\degree$ & {11.337$\degree$} & {11.636$\degree$} \\
            \ourName{} $\dagger$  & $6.264\degree$ & {6.818$\degree$} & $20.425\degree$ & $8.156\degree$ & $21.456\degree$ & {0.081$\si{m}$} & $21.057\degree$ & $12.474\degree$ & $14.029\degree$ \\
            A-SDF \cite{mu_a-sdf_2021} & \textbf{3.457$\degree$} & $30.740\degree$ & $7.189\degree$ & \textbf{3.884$\degree$} & $34.714\degree$ & $0.235\si{m}$ & $12.265\degree$ & $16.139\degree$ & $15.375\degree$ \\
            \ourName-No-Enf & $29.289\degree$ & $37.476\degree$ & $36.086\degree$ & $46.648\degree$ & $37.856\degree$ & \textbf{0.104$\si{m}$} & $34.451\degree$ & $35.892\degree$ & \makecell{$36.967\degree$} \\
            \ourName{} & $8.214\degree$ & \textbf{10.678$\degree$} & \textbf{6.815$\degree$} & $14.136\degree$ & \textbf{23.467$\degree$} & {0.141$\si{m}$} & \textbf{8.328$\degree$} & \textbf{11.512$\degree$} & \textbf{11.940$\degree$} \\
            \midrule
            A-SDF \cite{mu_a-sdf_2021} & \textbf{1.000} & \textbf{0.956} & 0.933 & \textbf{0.99} & \textbf{0.9} & \textbf{0.988} & 0.923 & \textbf{0.962} & \textbf{0.957} \\
            \ourName-No-Enf & 0.604 & 0.748 & 0.727 & {0.600} & 0.700 & 0.496 & 0.710 & 0.646 & 0.655 \\
            \ourName{} & {0.932} & {0.932} & \textbf{0.973} & 0.570 & {0.867} & {0.956} & \textbf{0.970} & {0.908} & {0.886} \\
            \bottomrule
        \end{tabular}
        \caption{Articulation State Prediction Results. We report the joint state error ($\downarrow$) in the first set of rows for all correctly classified joints and joint type accuracy ($\uparrow$) in the last set of rows. As A-SDF does not classify the joint type and \ourName{} trained on a single category always predicts the correct joint type, we do not report joint accuracy for those models. \textit{$^*$The joint state error mean is only reported across the revolute categories, as there is only one prismatic category.}}    
        \label{tab:joint_state_results}
    \end{subtable}
    \vspace{-0.3cm}
\end{table*}